\documentclass[journal]{IEEEtran}
 
\usepackage{balance}

\usepackage{booktabs}
\usepackage{multirow}
\usepackage{multicol}
\usepackage[normalem]{ulem}

\ifCLASSINFOpdf
  \usepackage[pdftex]{graphicx}
  \graphicspath{{../pdf/}{../jpeg/}}
  \DeclareGraphicsExtensions{.pdf,.jpeg,.png}
\else
  \usepackage[dvips]{graphicx}
  \graphicspath{{../eps/}}
  \DeclareGraphicsExtensions{.eps}
\fi
\usepackage{xcolor}

\usepackage{amsmath}
\usepackage{amsfonts}
\usepackage{bm}

\usepackage{subcaption}

\usepackage{url}

\begin{document}
%
\title{Towards properties of adversarial image perturbations \\
}

%
%
%

\author{
    \IEEEauthorblockN{Egor Kuznetsov\IEEEauthorrefmark{1}, Kirill Aistov\IEEEauthorrefmark{1}, Maxim Koroteev\IEEEauthorrefmark{1}\thanks{e-mail: koroteev.maxim@huawei.com}}
    \\\IEEEauthorblockA{\IEEEauthorrefmark{1} \small Huawei Luzin Research Center, Moscow, Russia
    \\}
}
%
%

\markboth{Huawei Technical Report}
{Koroteev \MakeLowercase{\textit{et al.}}:}
%



\maketitle
\thispagestyle{empty}

\begin{abstract}
Using stochastic gradient approach we study the properties of adversarial perturbations resulting in noticeable growth of VMAF image quality metric.
The structure of the perturbations is investigated depending on the acceptable PSNR values and based on the Fourier power spectrum computations for the perturbations. It is demonstrated that moderate variation of image brightness ($\sim 10$ pixel units in a restricted region of an image can result in VMAF growth by $\sim 60\%$). Unlike some other methods demonstrating similar VMAF growth, the subjective quality of an image remains almost unchanged. It is also shown that the adversarial perturbations may demonstrate approximately linear dependence of perturbation amplitudes on the image brightness. The perturbations are studied based on the direct VMAF optimization in PyTorch. The significant discrepancies between the metric values and subjective judgements are also demonstrated when image restoration from noise is carried out using the same direct VMAF optimization.
\end{abstract}

\begin{IEEEkeywords}
VMAF, video quality metrics, adversarial perturbations, adversarial examples
\end{IEEEkeywords}

%
\IEEEpeerreviewmaketitle

\section*{Introduction}
%
%
%
%

\IEEEPARstart{R}esearch conducted in the field of machine learning has demonstrated that neural networks under certain conditions are susceptible to adversarial perturbations \cite{SzegedyZSBEGF13} i.e., to the process of slight alteration of the model input (usually using gradient descent) in a way that turns out to be nearly imperceptible to humans but which may significantly change the predictions obtained with the models. This phenomenon represents a part of a more general one, namely vulnerability of neural networks to adversarial perturbations.
It is no wonder therefore that the most notable works in the adversarial machine learning field have been devoted to neural networks and the classification problem \cite{SzegedyZSBEGF13, GoodfellowSS14}, \cite{Kurakin}, \cite{CarliniWagner}. 

One of the important problems when analyzing the interaction of neural networks with adversarial perturbations is ability to generate these perturbations in the data. The majority of methods for producing adversarial examples rely on gradient-based optimization, however, some use other optimization techniques. Among them the authors of \cite{GoodfellowSS14} proposed the Fast Gradient Sign Method (FGSM) in which the adversarial perturbation is generated using the cross-entropy as a loss function and $l_{\infty}$ norm for measuring amplitudes of perturbations. The similar method called Fast Gradient Method (FGM) exploits the same loss function in combination with $l_{2}$ norm \cite{fgm}. In the current work we use the same gradient-based approach and investigate perturbations applying the same two norms. One important distinction of this work results from the use of different loss function. For this purpose we apply VMAF which, being a trainable machine learning model, also was shown to be vulnerable to adversarial perturbations. Unlike neural networks, in the case of image quality measuring models adversarial perturbations may not only be generated by some gradient-like method, but can also emerge as a result of image processing with already existing popular methods for image quality improvement. For example, in \cite{siniukov2021hacking} the authors apply various well-known image processing techniques, such as gamma correction, adaptive histogram equalizaion (CLAHE), unsharp masking and show that VMAF score can be significantly improved on average (by $\approx 60\%$). Many of these techniques give the highest gains when the image is substantially corrupted. This provides some evidence that the VMAF model diverges with the human estimate of image quality, indicating potential vulnerability of VMAF to adversarial perturbations. 

It is clear that various loss functions in principle can be used for the problem of adversarial perturbations analysis. The importance of exploiting VMAF is concerned with two reasons. First, this metric is widely used in the field of image/video processing for estimating the quality of image, especially in the industry. Second, we investigate adversarial perturbations resulting from the {\it direct optimization} of the VMAF metric and based on the implementation of this metric in PyTorch \cite{vmaf_reimplementation}, \cite{vmaf_torch}. In the latter work we analyzed how the stochasting gradient descent method behaves for the PyTorch implementation of VMAF and tried to argue that this metric does not result in singularities during the training process and is in good correspondence with the original VMAF. 

Thus, the goal of the present paper is twofold: 1) investigate and present properties of perturbations generated on images by means of direct VMAF optimization; 2) provide more evidence for applicability of the algorithm implementing VMAF proposed in \cite{vmaf_reimplementation}. For the latter goal we also consider the problem of image restoration from the point of view previously studied in   \cite{Ding_2021}, where various metrics were compared among themselves in terms of the quality of image restoration. Without pretending to noticeably new results {\it at this particular point}, we restore images using VMAF-torch numerically and demonstrate that our results are in correspondence with \cite{Ding_2021}. It may be worth noting at the outset that the question of the loss function properties wrt. adversarial perturbations is not studied in this paper and is assumed to be presented elsewhere: we only focus on the properties of the adversarial perturbations.





\section*{Adversarial perturbations}

Assume that a reference image R is represented as an element of a vector space $\mathbf{R}^{d}$, where each coordinate can take integer
values in the interval $[0, 255]$. Then we solve the following optimization problem
\begin{equation}
\delta^{*} = \underset{\|\delta\|\leq \varepsilon}{\operatorname{argmax}} \operatorname{VMAF}(R, R+\delta),
\label{maximization}
\end{equation}
where $\|\cdot\|$ is some norm on a vector space and $\delta\in\mathbf{R}^{d}$ may be referred to as {\it perturbation}. 
 Common choices for the norm in the above expression are $l_{\infty}$ norm and $l_{2}$ norm; we generate perturbations for both. It may be worth stressing from the beginning that in this formulation the problem implies that the signal $R$ is perturbed {\it additively}, and thus the problem can be thought of as that of a quality metric maximization by means of an additive component, which, in a sense, plays the role of a noise, in the same way as the task of signal-to-noise ratio (SNR) maximization in the presence of gaussian additive noise can be formulated. Another alternative to this formulation would be application of a filter to the original signal $R$, i.e., a perturbation which is obtained by convolving the original signal with some other signal instead of addition. 

The numerical solution of this problem may be affected by those  elements of $\mathbf{R}^{n}$, over which we search for the extremum. We use images extracted from video streams publicly available at \cite{netflix}.  The use of this data is justified, as it is the very data that the default version of VMAF was designed for. Concerning the VMAF algorithm  handling, we disable the final score clipping to $[0, 100]$ range.

\subsection{$l_{\infty}$ norm}

First, we use $l_{\infty}$ norm: for $v\in\mathbf{R}^{d}$, where $v=\{v_{i}\}_{i=1}^{d}$ we take $\|v\|_{\infty}=\max_i|v_i|$. 
Note that setting $\varepsilon$ in (\ref{maximization}) to some specific value $\varepsilon^{*}$ in this case means that each pixel of $R$ 
differs from corresponding pixel of $R+\delta$ by $\varepsilon^{*}$ units at most.

We use the standard projected gradient descent (PGD) method to generate adversarial examples. In PGD the single update can be written for our case as
$$
\delta=\mathcal{P}\left(\delta+\alpha \nabla_\delta \operatorname{VMAF}(R, R+\delta) \right),
$$
where $\mathcal{P}$ denotes the projection onto the ball of radius $\varepsilon$. For the case of $l_{\infty}$ norm
this takes the form
$$
\mathcal{P}(x) = \operatorname{clip}(x, \operatorname{min}=-\varepsilon, \operatorname{max}=\varepsilon),
$$
where $\operatorname{clip}(x,\operatorname{min},\operatorname{max})$ ensures that the perturbed values stay within the allowed range. 
The updates are repeated until convergence.
For the computations, we set $\alpha=0.1$ and use the number of steps from $50$ to $150$ depending on $\varepsilon$.

\begin{figure}
	\centering
	\includegraphics[width=1.0\linewidth]{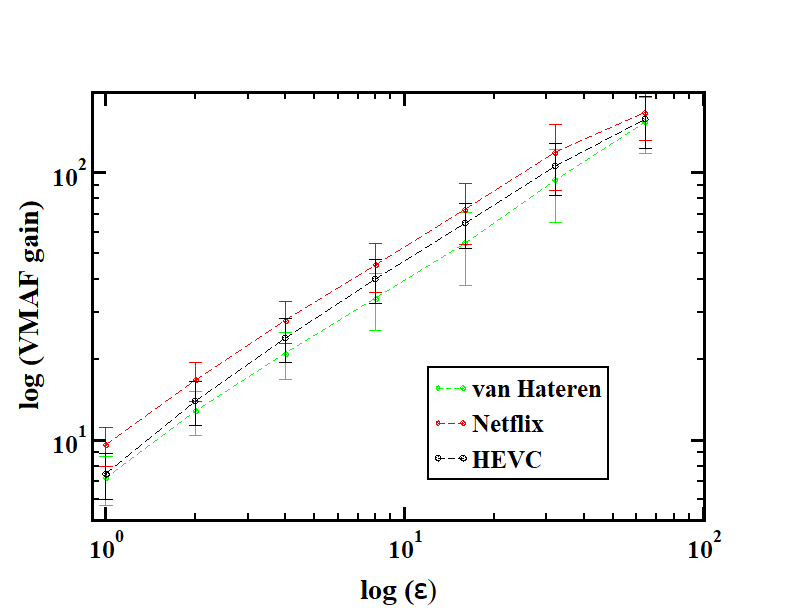}
	\caption{Average VMAF gain (log-log scale) achieved for perturbations of various amplitudes $\epsilon$ shown on x-axis. Three data sets were used: two sets of video streams, Netflix \cite{netflix} and HEVC; $100$ frames were collected from each; one image data set was used, due to van Hateren \cite{vanHateren}, $100$ images were randomly chosen from it. }
	\label{fig:vmaf_gain}
\end{figure}

\begin{figure*}
	\centering
	\begin{subfigure}[b]{0.475\textwidth}
		\centering
		\includegraphics[width=\textwidth]{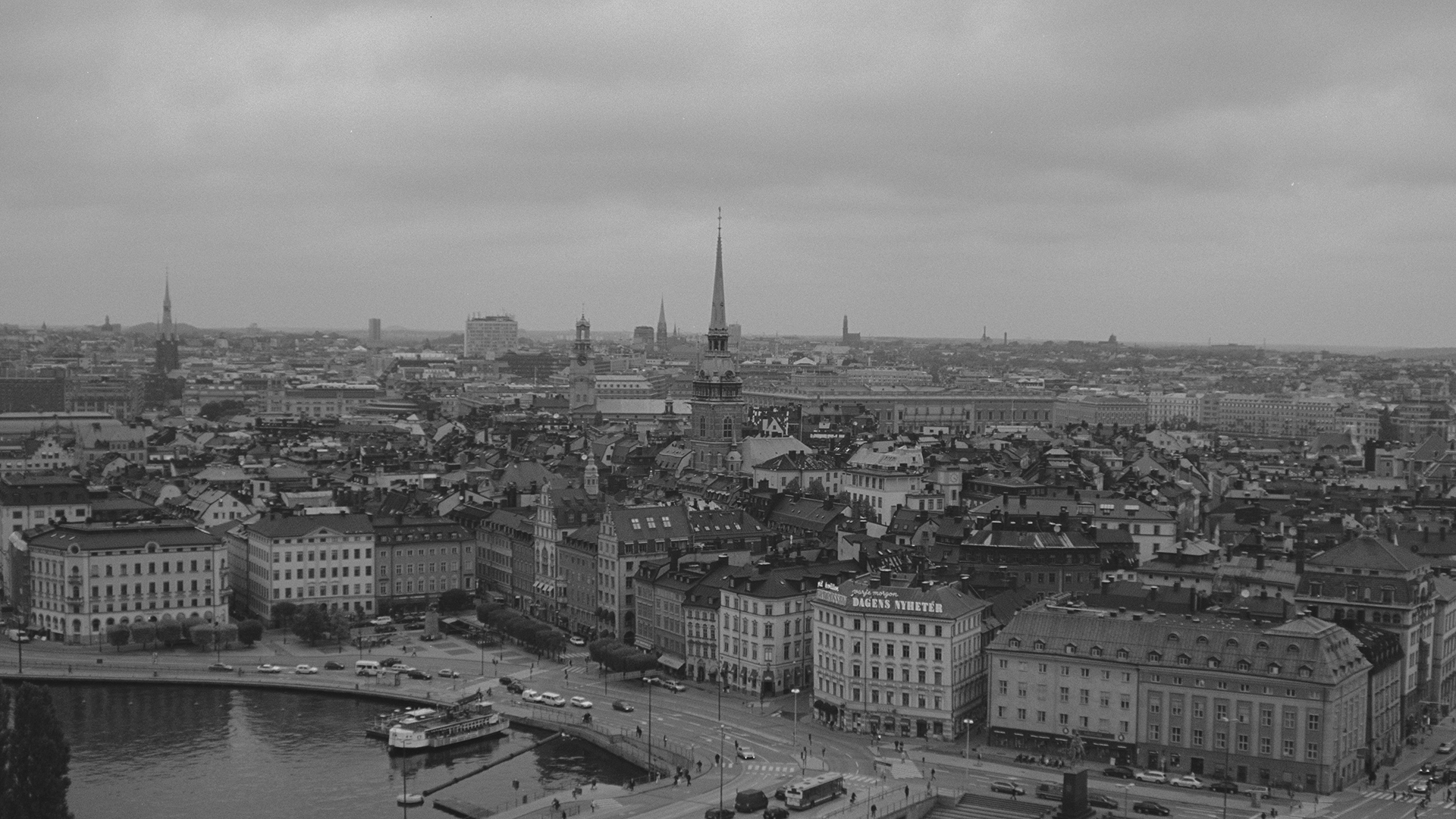}
		\caption{Reference image}    
		\label{fig:old_town_cross_reference}
	\end{subfigure}
	\hfill
	\begin{subfigure}[b]{0.475\textwidth}  
		\centering 
		\includegraphics[width=\textwidth]{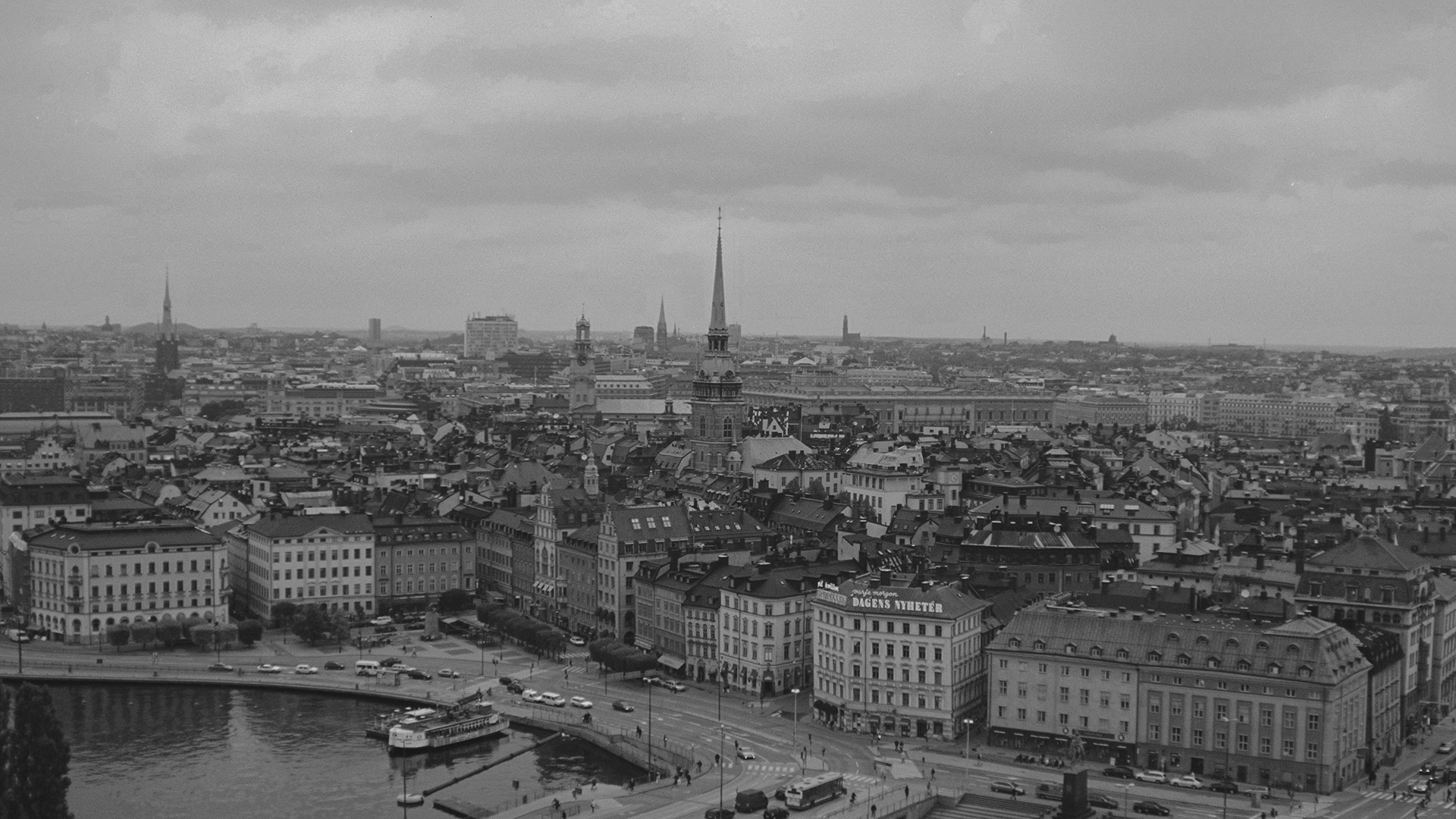}
		\caption{Perturbed image $R+\delta$. }   
		\label{fig:old_town_cross_processed}
	\end{subfigure}
	\vskip\baselineskip
	\begin{subfigure}[b]{.65\textwidth}   
		\centering 
		\includegraphics[width=\textwidth]{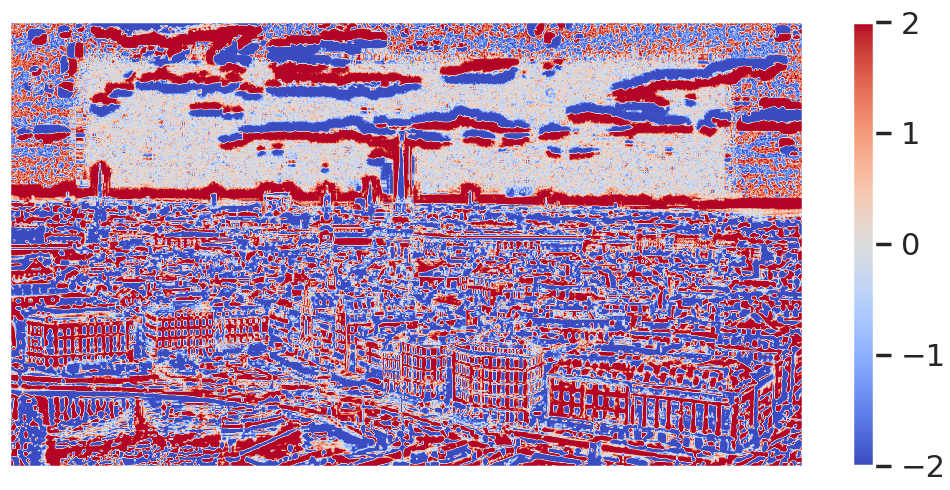}
		\caption{Perturbation image}    
		\label{fig:old_town_cross_delta}
	\end{subfigure}
	\caption{Adversarially perturbed image for $l_\infty$ norm used in (\ref{maximization}), $\varepsilon=2$,  $\operatorname{VMAF}(R, R+\delta)=112.5$ } 
	\label{fig:old_town_cross_attack}
\end{figure*}

The resulting {\it gain} computed as $\operatorname{VMAF}(R, R+\delta) - \operatorname{VMAF}(R, R)$ for the adversarial perturbation $\delta$ obtained after sufficient number of steps in PGD and averaged over all images in the dataset is shown in Fig. \ref{fig:vmaf_gain} for different $\varepsilon$. 
It can be seen that perturbing pixels of the image with $\delta$ by as little as one pixel value can result in VMAF growth up to $10$ units. It is also noticeable that the dependence of the gain on the perturbation amplitude is approximately power-law with the exponent $\approx 2.2$, though we do not have additional evidence for any power-law behavior here. On the other hand, Fig. \ref{fig:vmaf_gain} shows that the results on the VMAF growth are hardly dependent on the data set for different  $\varepsilon$: all three data sets demonstrate very close behavior. The perturbed image $R+\delta$ along with the reference $R$ and the perturbation $\delta$ image are shown in Fig. \ref{fig:old_town_cross_attack}.

%
%
%

Both figures show that significant VMAF increase can be achieved even for relatively small maximum perturbation amplitudes $\delta$ and in the same time visually this perturbation is hardly perceptible. The fact that some small perturbations $\epsilon \ll \bar{p}$ may result in significant VMAF growth is a known peculiarity of this metric.

\subsection{$l_{2}$ norm}

We also search the solution of the above optimization problem with respect to $l_{2}$ norm 
$\|v\|_2 = \sqrt{v_1^2 + \cdots + v_n^2}$.
This norm is related to another widespread quality assessment metric, PSNR.
Given a reference image $R$ with resolution $m \times n$, $R \in \mathbf{R}^d, d=mn$ and some perturbation $\delta$, we can write
$$
\operatorname{PSNR}(R, R+\delta) = 10 \log\left(\frac{255^2}{\operatorname{MSE}(R, R+\delta)}\right)= 
$$
$$
= 10 \log\left(\frac{255^2}{\frac{1}{mn}\|R - (R+\delta)\|^{2}_2}\right) = 10 \log\left(\frac{255^2 mn}{\|\delta\|^{2}_2}\right),
$$
where MSE stands for the mean squared error.
Thus, $\operatorname{PSNR}(R, R+\delta)$ does not depend on $R$, but only on the perturbation norm value $\|\delta\|_2$,
If $\|\delta\|_2 \leq \varepsilon$, then
$$
\operatorname{PSNR}(R, R+\delta) =  10 \log\left(\frac{255^2 mn}{\|\delta\|^{2}_2}\right) \geq 10 \log\left(\frac{255^2 mn}{\varepsilon^{2}}\right),
$$
so setting a certain value for $\varepsilon$ effectively sets a minimum PSNR value between the reference and preprocessed image.

We can choose $\varepsilon$ to achieve certain lower bounds $p$ for PSNR by setting
\begin{equation}
\varepsilon=255\sqrt{mn} \cdot 10^{-\frac{PSNR}{20}}
\label{epsilon_psnr}
\end{equation}
which comes from the inversion of the formula above. The results for VMAF gains taking into account the boundary for PSNR are shown in Fig. \ref{fig:vmaf_vs_psnr_l2}. It can be noticed that both norms are able to provide comparable growth in terms of VMAF for some reasonable range of perturbation amplitudes. 

\section*{Analysis of perturbations}

In the previous section we solved the optimization task by maximizing VMAF image quality metric and computing the amplitudes of perturbations resulting in the optimal solution. In this section we would like to look at the properties of the perturbations more closely.

%
\begin{figure}
	\centering
	\includegraphics[width=1.0\linewidth]{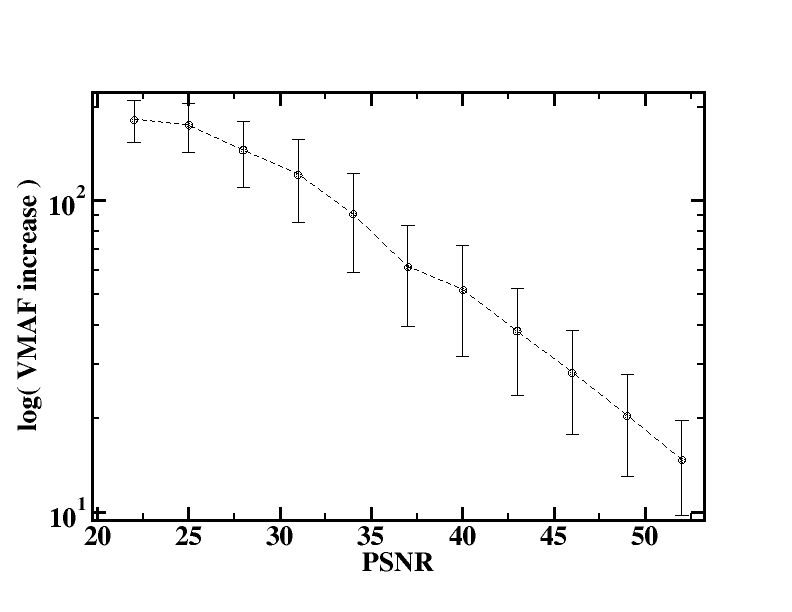}
	\caption{Average VMAF gain (log scale) wrt. $l_{2}$ norm achieved for perturbations of various amplitudes with MSE corresponding to PSNR values shown on x-axis. Dataset: netflix, $10$ images were collected from each stream ($90$ images in total).}
	\label{fig:vmaf_vs_psnr_l2}
\end{figure}
\begin{figure*}
	\centering
	\begin{subfigure}[b]{0.475\textwidth}
		\centering
		\includegraphics[width=\textwidth]{figures/30_OldTownCross_25fps_reference.png}
		\caption{Reference image}    
		\label{fig:old_town_cross_reference_l_2}
	\end{subfigure}
	\hfill
	\begin{subfigure}[b]{0.475\textwidth}  
		\centering 
		\includegraphics[width=\textwidth]{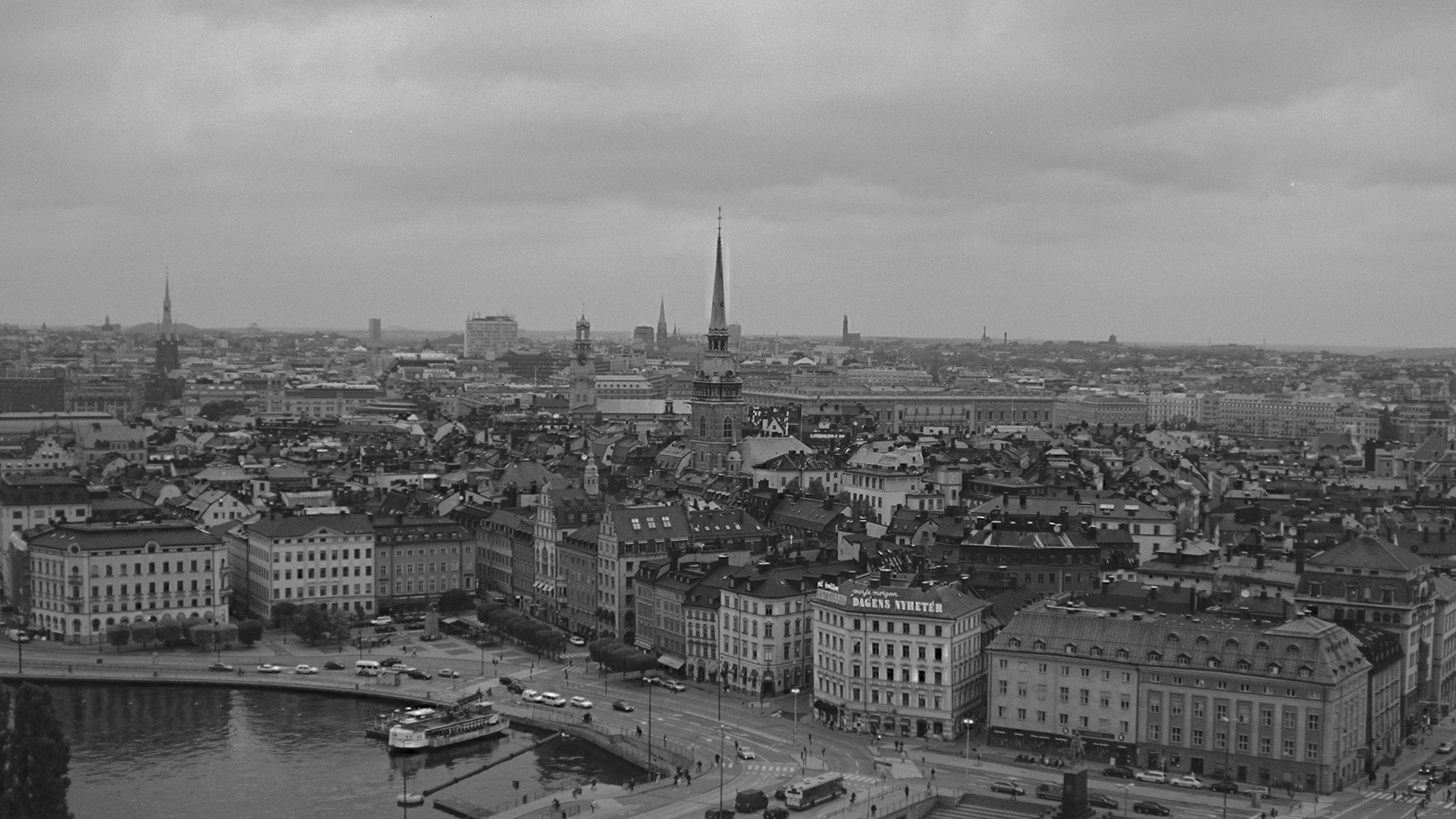}
		\caption{Perturbed image $R+\delta$}   
		\label{fig:old_town_cross_processed_l_2}
	\end{subfigure}
	\vskip\baselineskip
	\begin{subfigure}[b]{.65\textwidth}   
		\centering 
		\includegraphics[width=\textwidth]{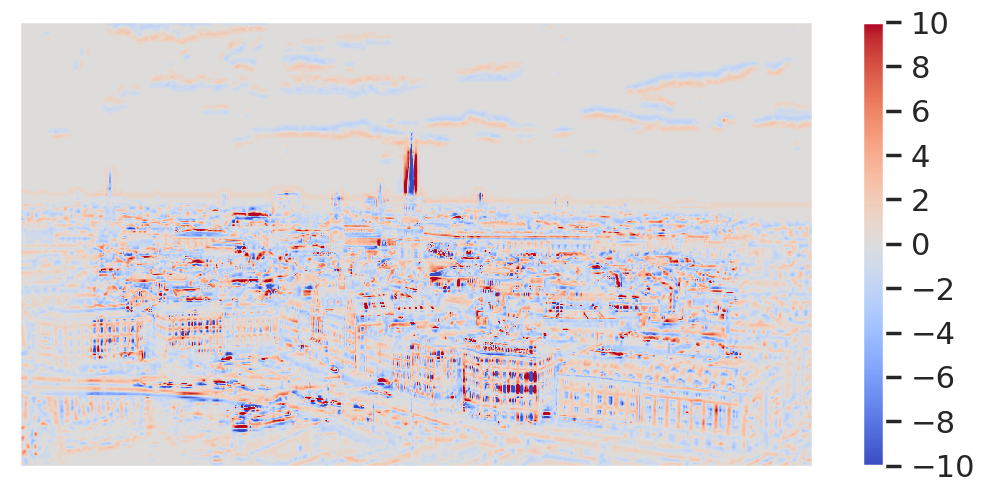}
		\caption{Perturbation image}    
		\label{fig:old_town_cross_delta_l_2}
	\end{subfigure}
	\caption{Adversarially perturbed image for $l_2$ norm, $\operatorname{PSNR}(R, R+\delta)=40$,  $\operatorname{VMAF}(R, R+\delta)=134.1$ } 
	\label{fig:old_town_cross_attack_l_2}
\end{figure*}

It can be seen in Fig. \ref{fig:vmaf_vs_psnr_l2} that the decrease of PSNR for the adversarial perturbations results in increasing VMAF gain. Note that the decrease of PSNR implies a smaller signal to noise ratio and, consequently, the growth of the noise (distortions) in the system. Generally speaking, this should lead to a decline in image quality and to the decrease of VMAF which is thought to be correlated to the subjective human judgement about image quality, but the results demonstrate the inverse behavior. It is clear that, on the one hand, this is related, in some way, to the specific properties of VMAF used as a cost function and optimiztion over it. More specifically this may imply that the effective noise, increasing in the system in this case can not be thought of as a generic form of noise and represents a very specific noise-like perturbation. Note, that this situation is not unique: when applying enhancement filters to images the situation may be similar. Sharpening of edges infers distortions into the signal, so that PSNR would drop, but in the same time VMAF may increase. However, even said that the perturbation in this case is noise-like we did not imply it was completely random. Its non-randomness is clearly seen in Figs. \ref{fig:old_town_cross_attack} and \ref{fig:old_town_cross_attack_l_2}, where the reference and perturbed images are shown alongside the computed perturbation $\delta$. 
The latter demonstrates a lot of structure resembling the reference image. 

\begin{figure}
	\centering
	\includegraphics[width=1.0\linewidth]{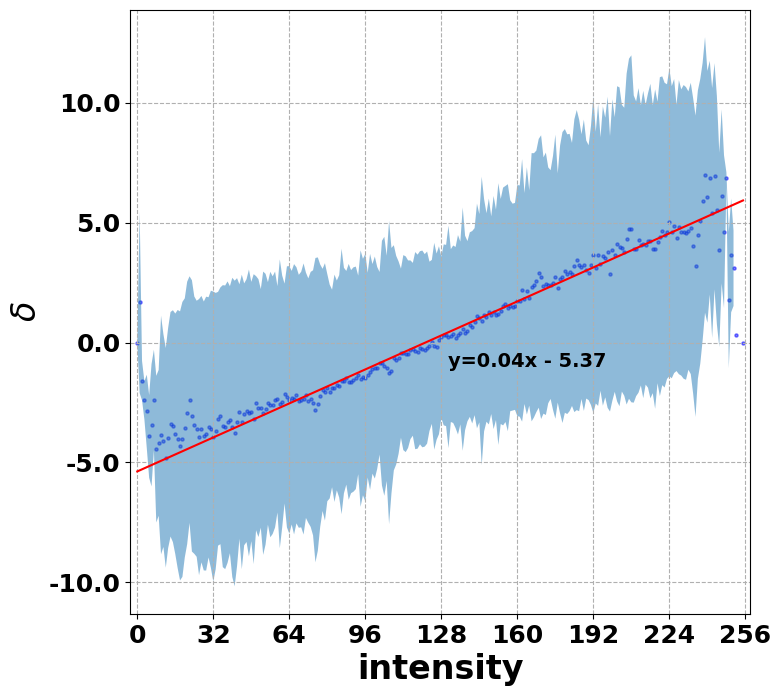}
	\caption{The dependence of the intensity perturbation on the intensity. The perturbation $\delta$ was obtained by the learning procedure explained in the main text. For each pixel of the image the magnitude of its intensity and the corresponding perturbation in this pixel were plotted as an $(x,y)$ pair. For each intensity the perturbations were averaged to obtain the mean value of the perturbation for the given intensity (shown by the blue points). The figure shows the results for $40$ images and the light blue band shows the standard deviations. The upper bound for perturbations was chosen as $PSNR\ge 40$. The perturbations were obtained using $l_{2}$ norm.}
	\label{fig:delta_vs_brightness}
\end{figure}

To give another characterization of this perturbation, we can examine its dependence on the brightness of the pixel. The resulting relations are close to linear, so that brighter pixels on average have larger magnitudes of perturbations and vice versa (Fig. \ref{fig:delta_vs_brightness}). 
This behavior obviously differs from addition of independent noise to the signal and can be accounted for if we notice that the increase of VMAF metric generically is caused by some form of contrasting. This means that the edges and boundaries of the objects in the image become sharper. As these boundaries are usually represented as regions of sharp changes of the brightness, the sharpening means that generically bright pixels have to become even brighter, while dark pixels become darker, which is in the correspondence to our observations. However, the linear dependence we observe is interesting. It becomes more distinct if we take into account not all pixels but only those which are located near edges and boundaries, i.e., in the regions of abrupt brightness variation. These pixels can be identified by various methods. We apply the Lapalce operator in each pixel and compare the magnitude of the laplacian with the threshold determined by the variance of the image. The resulting linear dependence shown in Fig. \ref{fig:delta_vs_brightness} seems to be quite certain despite the small slope. On the other hand, it is clear that we can not expect a significant slope in this experiment just because maximization of VMAF implies that the perturbations of the image can not be too strong as this inevitably would result in image quality degradation. In addition, the perturbations $\delta$ in Fig. \ref{fig:delta_vs_brightness} are controlled by the condition $PSNR\ge 40$.

\begin{figure}
	\centering
	\includegraphics[width=1.0\linewidth]{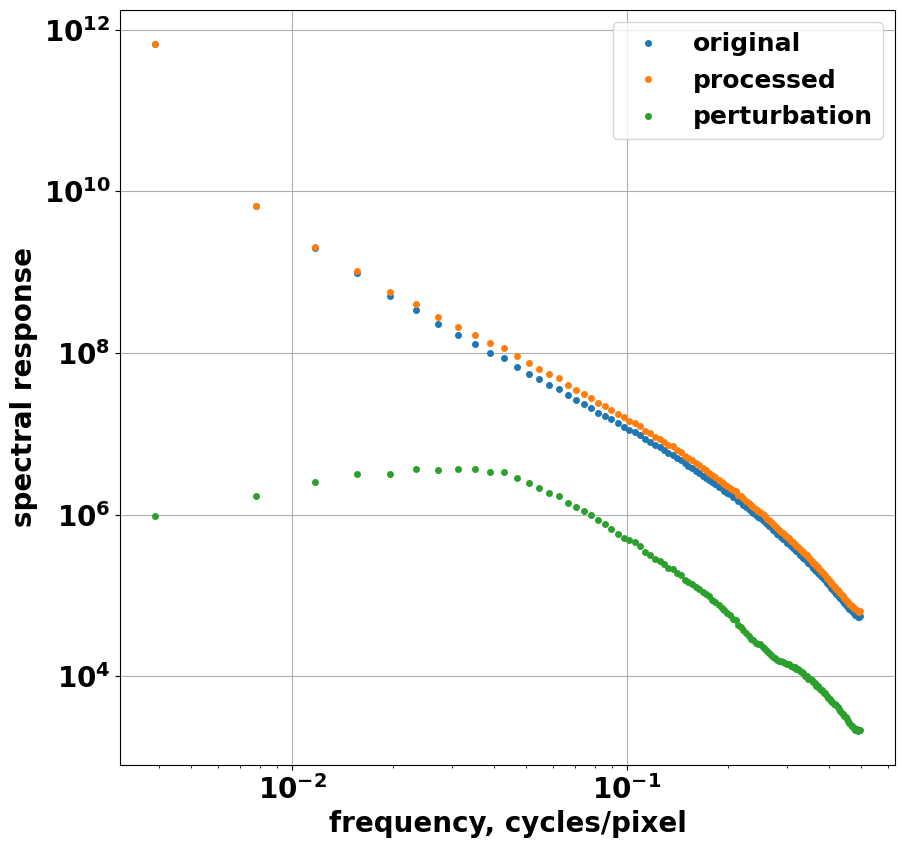}
	\caption{1D power spectra for the reference, perturbed, and perturbation image. The spectra were obtained by, first, collecting $100$ random patches from the images; the size of the patch was $256\times256$ pixels. Then 2D spectrum was computed for each patch. The 2D spectra were then averaged and finally an additional circular averaging was applied to obtain the final 1D spectrum.}
	\label{fig:spectra}
\end{figure}

The characterization of the perturbations can also be given by portreying 1D power spectra for the original, perturbed, and perturbation images, which are presented in Fig. \ref{fig:spectra}\footnote{The method for obtaining these spectra is described in \cite{Ruderman}. We also used it in \cite{Koroteev2021}. Note, that portraying 2D spectra in this case is not quite indicative.}.
Overall it is seen that the perturbation of the {\it image spectrum} magnitude produced by the adversarial perturbation in the pixel domain is insignificant, which is also seen from the amplitude of the spectrum for the perturbation which is at least two orders of magnitude smaller than the reference and perturbed spectra (Fig. \ref{fig:spectra}). The spectrum for the perturbation image may account for the observations in Fig. \ref{fig:old_town_cross_attack_l_2}c. It is seen that the tails of the spectra both for the reference and perturbation images exhibit similar behavior, keeping the mutual structure of frequencies. This corresponds to smaller objects, edges, lines etc. observed in Figs. \ref{fig:old_town_cross_attack_l_2}, thereby reproducing the objects in the reference image. On the other hand, the low frequency part of the spectrum is noticeably reduced for the perturbation signal, which corresponds to disappearance of larger structures compared to the reference image. Overall, the behavior of the spectra is close to linear, which is in good correspondence to the classical results concerning the spectra for natural images \cite{Ruderman}. At the same time the structure of the spectra shown in Fig. \ref{fig:spectra} implies the corresponding correlation function incompatible to the independent noise, which provides an additional evidence of the complex nature of the perturbations.

Despite these observations the computations also show that 1D power spectrum for a sample of images is weakly sensitive to VMAF growth (cf. blue and red curves in Fig. \ref{fig:spectra}). Realistic VMAF variations resulting from  standard enhancement methods may provide improvements $\sim 4-6$ VMAF units, while for the perturbed image in Fig. \ref{fig:old_town_cross_attack_l_2} VMAF attains the magnitude $134$, significant increase, taking into account that the normal range for high-quality images is $\sim 85-95$ VMAF units. This significant increase, however, is hardly can be viewed in the spectra in Fig \ref{fig:spectra}.

\begin{figure}
	\centering
	\includegraphics[width=1.0\linewidth]{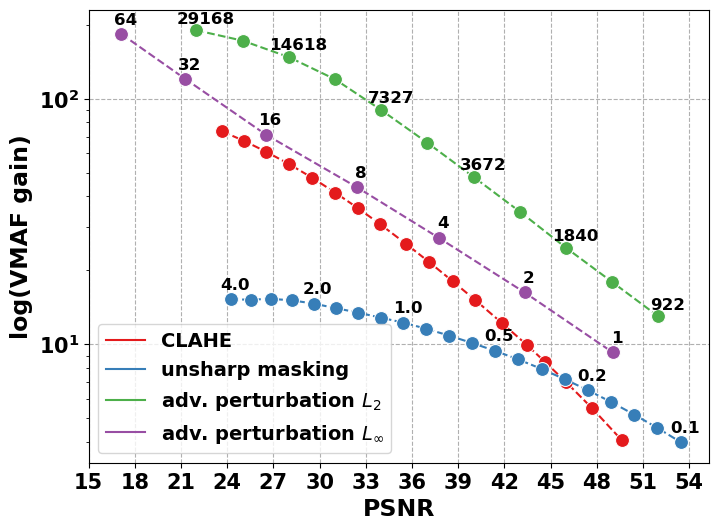}
	\caption{Average VMAF gain achieved by the unsharp masking filter (size $5\times 5$), CLAHE, and adversarial perturbations using two norms $l_{\infty}$ and $l_{2}$ on Netflix dataset for comparable PSNR values. Different points on the red and blue curves represent different values of amplification parameters for  CLAHE and the unsharp masking correspondingly; we set these parameters so that the processed images fall withing PSNR range of $40-50$ dB. The points on the green curve correspond to various PSNR lower bounds established when training the perturbation model. The points on the violet curve correspond to the maximum perturbation magnitudes $\varepsilon$, shown with the numbers near each point.}
	\label{fig:attack_vs_unsharp}
\end{figure}

\begin{figure*}
	\centering
	\begin{subfigure}[b]{0.475\textwidth}
		\centering
		\includegraphics[width=\textwidth]{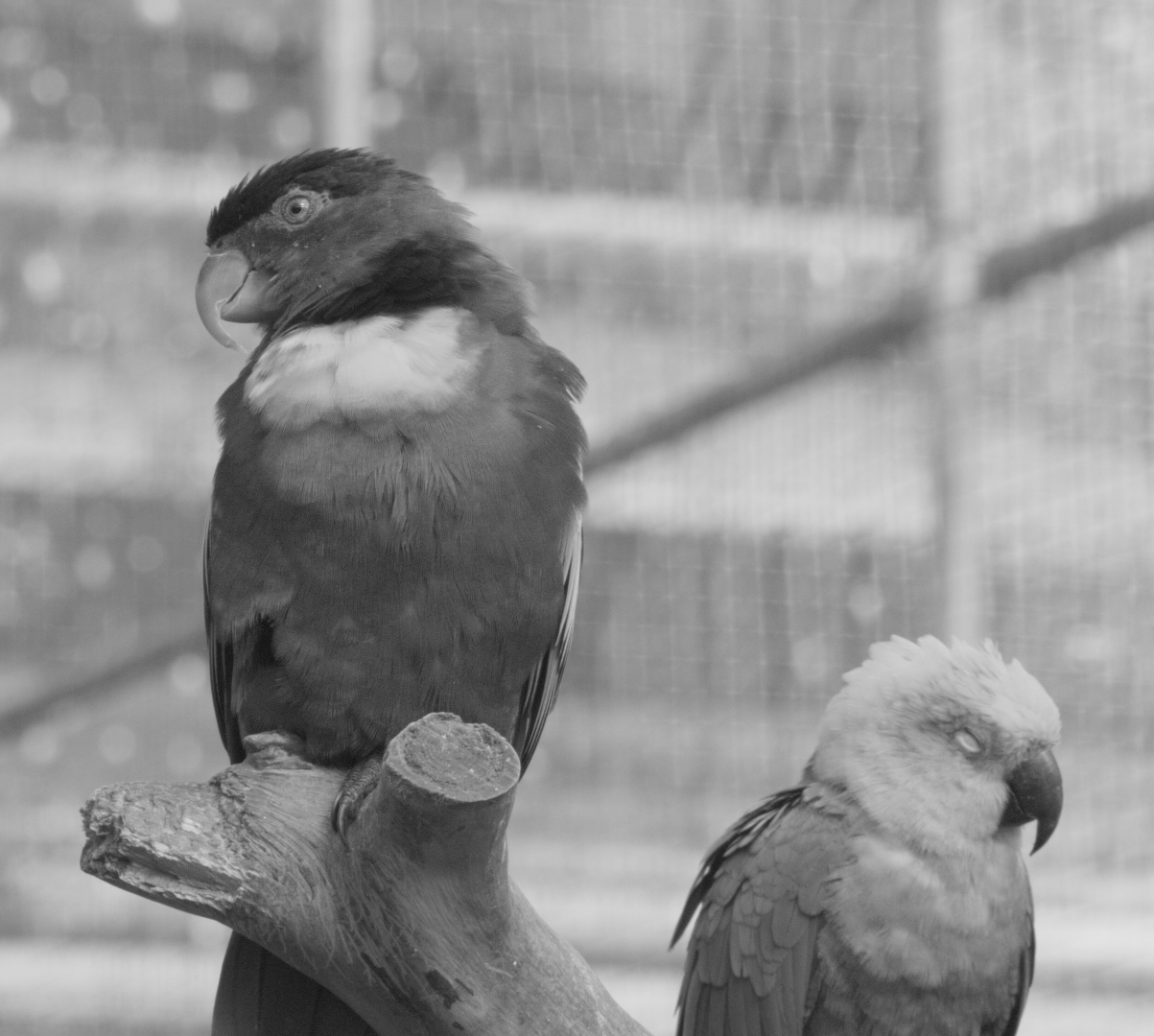}
		\caption{Reference image}    
	\end{subfigure}
	\hfill
	\begin{subfigure}[b]{0.475\textwidth}  
		\centering 
		\includegraphics[width=\textwidth]{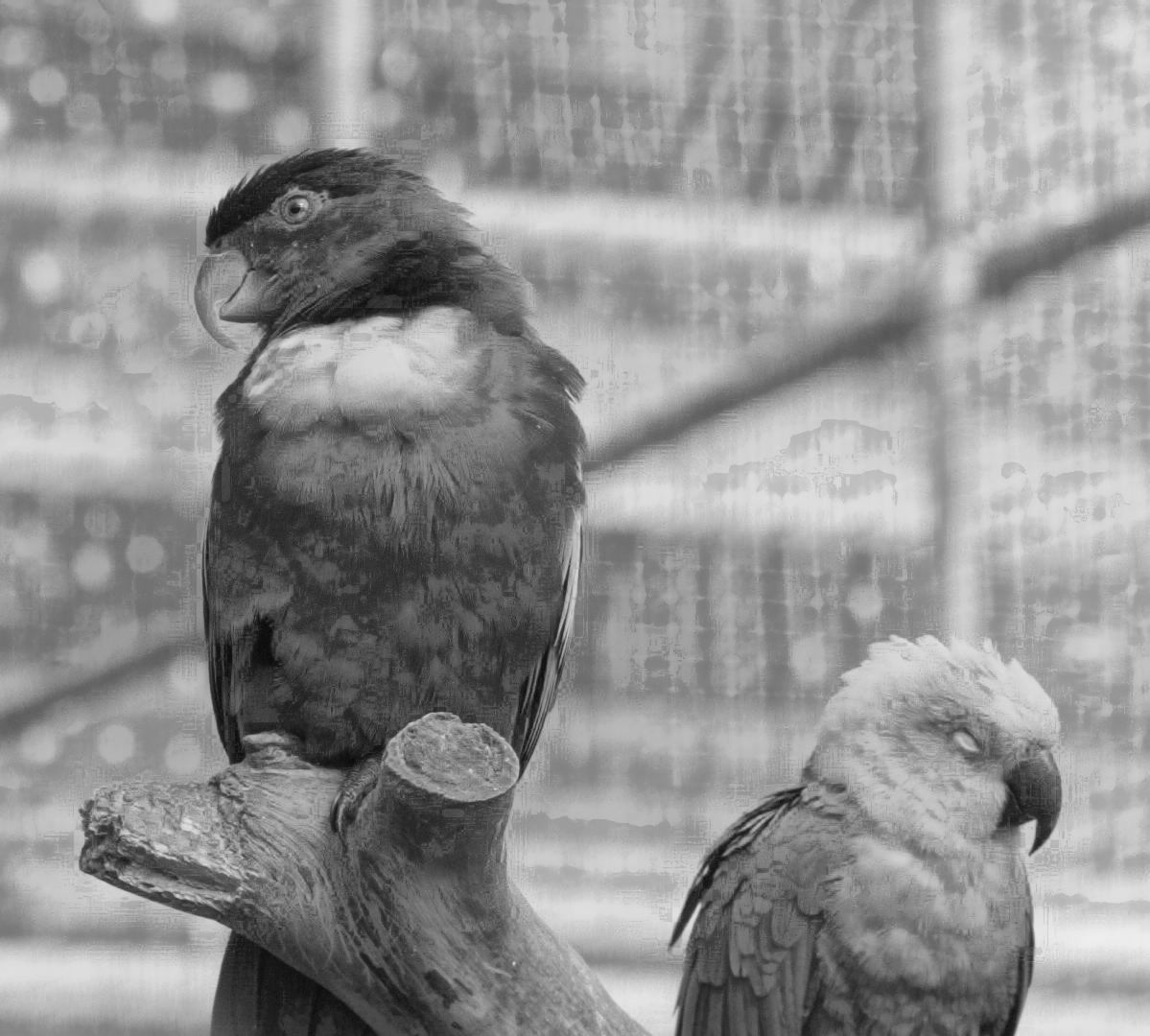}
		\caption{$l_{\infty}$ perturbed image}   
	\end{subfigure}
	\vskip\baselineskip
	\begin{subfigure}[b]{0.475\textwidth}   
		\centering 
		\includegraphics[width=\textwidth]{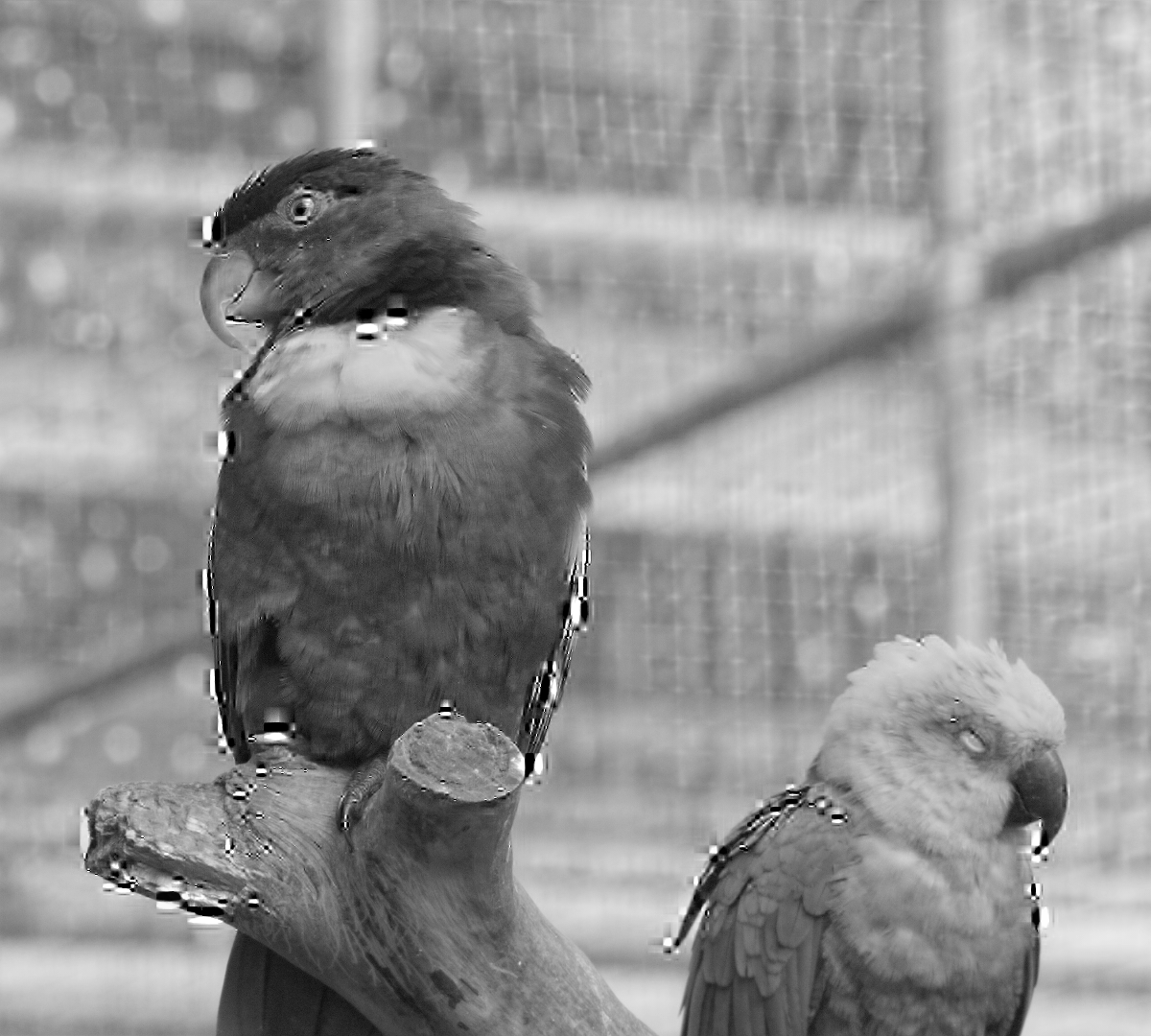}
		\caption{$l_{2}$ perturbed image.}    
	\end{subfigure}
	\hfill
	\begin{subfigure}[b]{0.475\textwidth}   
		\centering 
		\includegraphics[width=\textwidth]{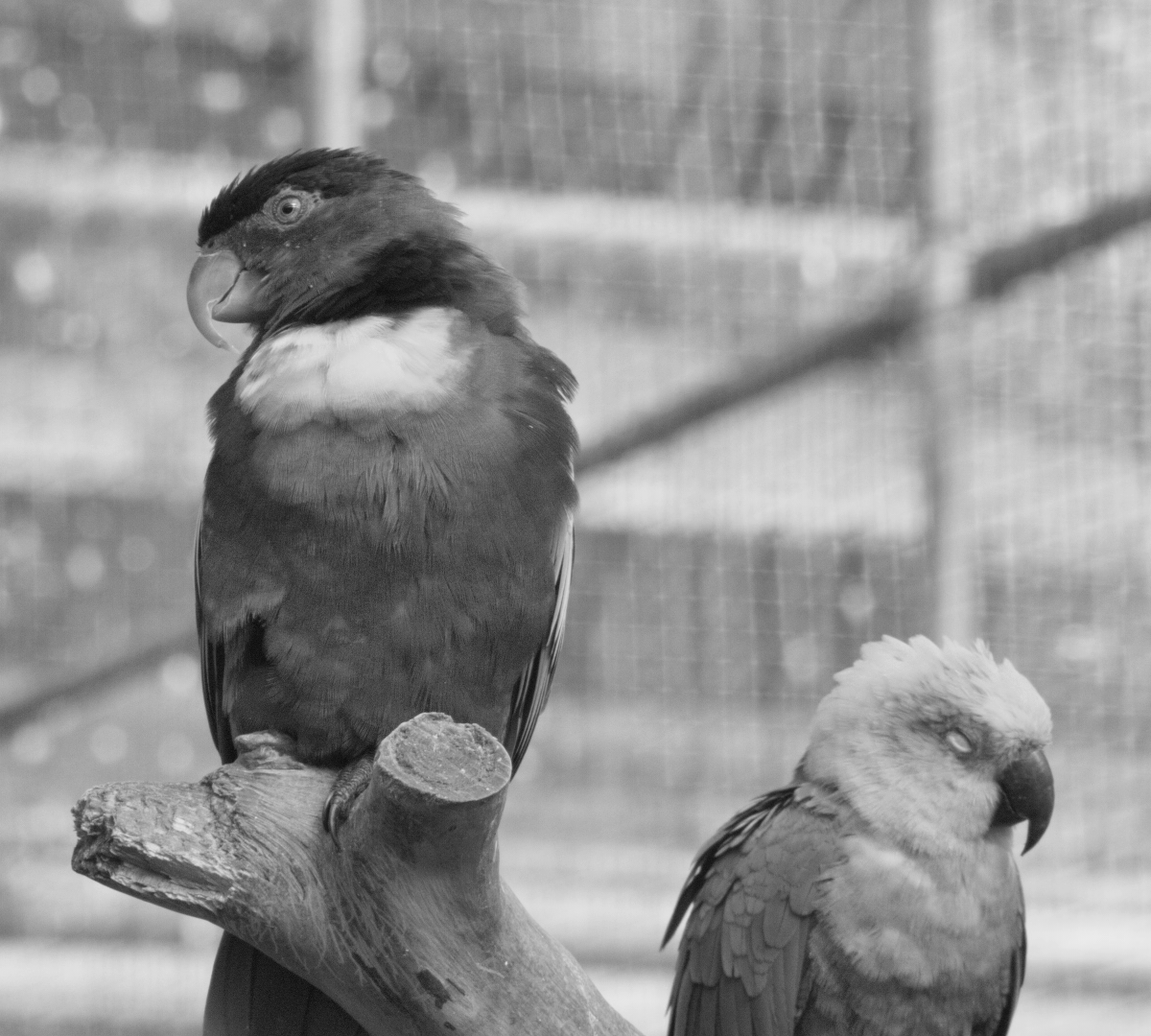}  
		\caption{CLAHE}
	\end{subfigure}
	\vskip\baselineskip
	\hfill
	\caption{Comparison of images enhanced with the adversarial perturbations and CLAHE. The images approximately correspond to the results of Fig. \ref{fig:attack_vs_unsharp} for PSNR values $30$. For the perturbation amplitudes this PSNR corresponds to $\epsilon\approx 10$ and $\epsilon\sim 10^{4}$ for the norms $l_{\infty}$ and $l_{2}$ correspondingly. The latter difference accounts for large VMAF values obtained for the latter norm and noticeable artefacts seen in the image (c).} 
	\label{fig:various_methods_enhancement}
\end{figure*}

Some classical image processing methods like unsharp masking and contrast limited adaptive histogram equalization (CLAHE) are also known to enhance VMAF score \cite{siniukov2021hacking}. It is instructive then to consider them together with the adversarial perturbations. We compare VMAF enhancement wrt PSNR achieved by means of adversarial perturbations to these methods. The plot is shown in Fig. \ref{fig:attack_vs_unsharp}. 
It can be seen that VMAF growth achieved by two types of adversarial perturbations we constructed is significantly higher than that attained with the methods like CLAHE and unsharp masking. Note, that for small PSNR values $\sim 20$ the growth compared to the unsharp masking can attain the order of magnitude. It should be noted, however, that such small values of PSNR are not realistic and virtually useless from the practical point of view, but even for PSNR $\sim30-35$ the growth is significant. Visual comparison of the enhancement achieved with these methods for PSNR values $\sim 30$ are shown in Fig. \ref{fig:various_methods_enhancement}. This figure shows some interesting peculiarities. For $l_{2}$ norm perturbations, for which the growth of VMAF is the largest noticeable artefacts appear on the image despite huge VMAF values. The character of these artefacts indicates they originate from ADM submetrics which turns out to be insensitive to this kind of distortions. Another thing to note in Fig. \ref{fig:attack_vs_unsharp} is that the behavior of perturbations turns out to be close to the straight line. For $l_{\infty}$ norm this regime is even extended into the region of small PSNR values. It is also worth noting that PSNR is proportional to log of MSE and hence VMAF gain for the perturbations invloved has a roughly power-law dependence on MSE.

We find that adversarial processing tends to enhance edges and change local contrast, which is not surprising since VMAF is known to prefer images with sharp edges. This behavior is parallel in a sense to the observations obtained for typical adversarial examples for neural networks which to human eye tend to look like artifacts or random noise. 

\section*{Reference image recovery}

The discussion concerning the perturbations and the corresponding VMAF growth presented in the previous sections was quite generic. To try to investigate the perturbations in connection with some more practical task of image processing we address the problem of image recovery from noise. One of the goals pursued here is a comparison with the results obtained in \cite{Ding_2021}, where a number of metrics were used for the same purpose but in our approach we can {\it directly} optimize over VMAF using the implementation of VMAF on Pytorch which we presented recently\cite{vmaf_reimplementation}\footnote{The implementation is available opensource at \cite{vmaf_torch}.}. Note that other tasks studied in \cite{Ding_2021}, such as denoising or deblurring and different FR metrics ranking are not considered here; we restrict ourselves with the analysis of image restoration problem using VMAF implementation.
 
The problem formulation is as follows. Given a reference image R and an initial image $I_0$ we try to recover R from $I_{0}$ by finding 
$$
I^{*}=\operatorname{argmax}_I \operatorname{M}(R,I), 
$$ 
where $M$ is the target full-reference metric.
For example, for PSNR the trivial solution is $I^*=R$, since it minimizes the MSE which PSNR is an inversely proportional function of.

\begin{figure}
	\centering
	\includegraphics[width=1.1\linewidth]{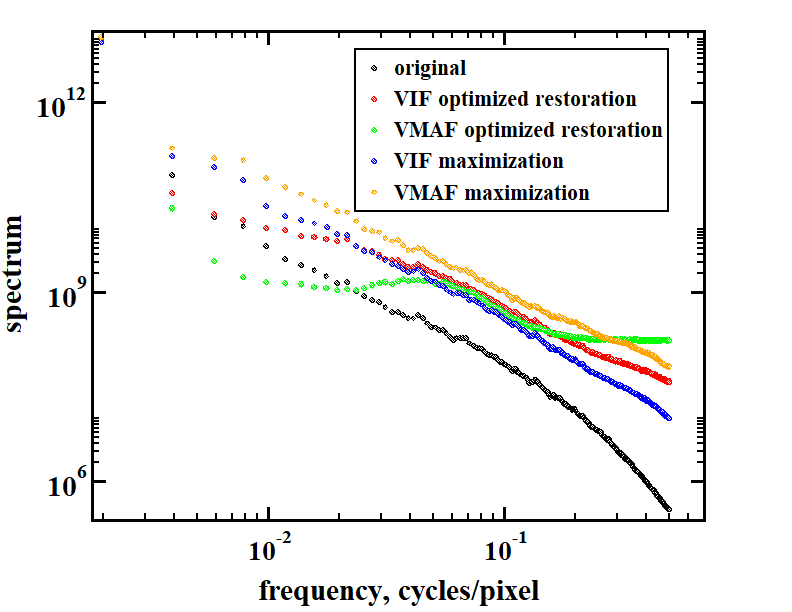}
	\caption{1D power spectra for the reference and restored images obtained by $VIF_{0}$ and VMAF optimizations and corresponding to Fig. \ref{fig:images_from_noise} a), d), h) correspondingly. In addition, 1D power spectra obtained for the images obtained by $VIF_{0}$ and VMAF metrics maximization and corresponding to Fig. \ref{fig:images_from_noise_unclipped} d), h) are shown. The spectra were computed by the method explained in the caption to Fig. \ref{fig:spectra}.}
	\label{fig:restoration_spectra}
\end{figure}

\begin{figure*}
	\centering
	\begin{subfigure}[b]{0.475\textwidth}
		\centering
		\includegraphics[width=\textwidth]{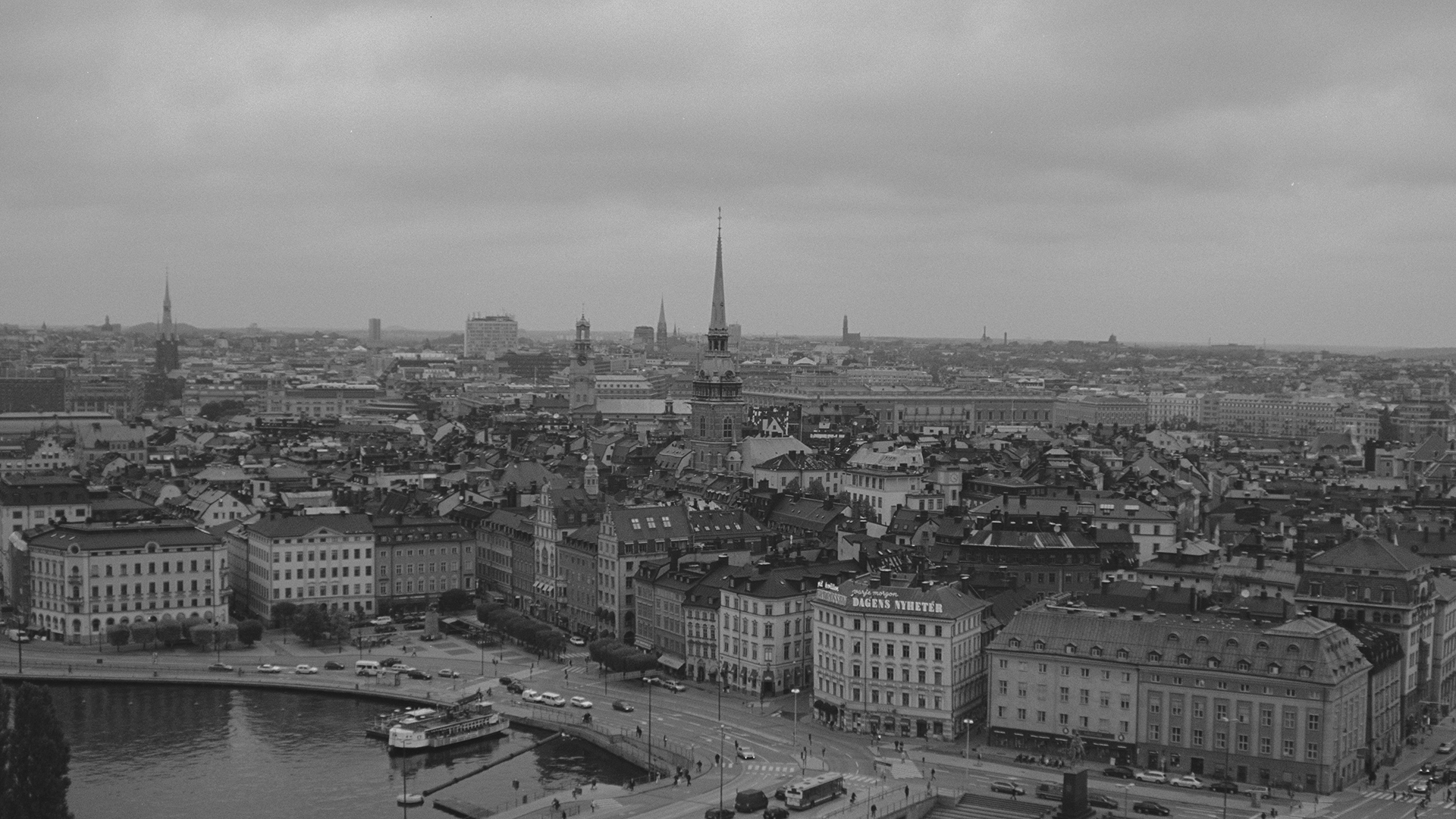}
		\caption{Reference image}    
		\label{fig:reference}
	\end{subfigure}
	\hfill
	\begin{subfigure}[b]{0.475\textwidth}  
		\centering 
		\includegraphics[width=\textwidth]{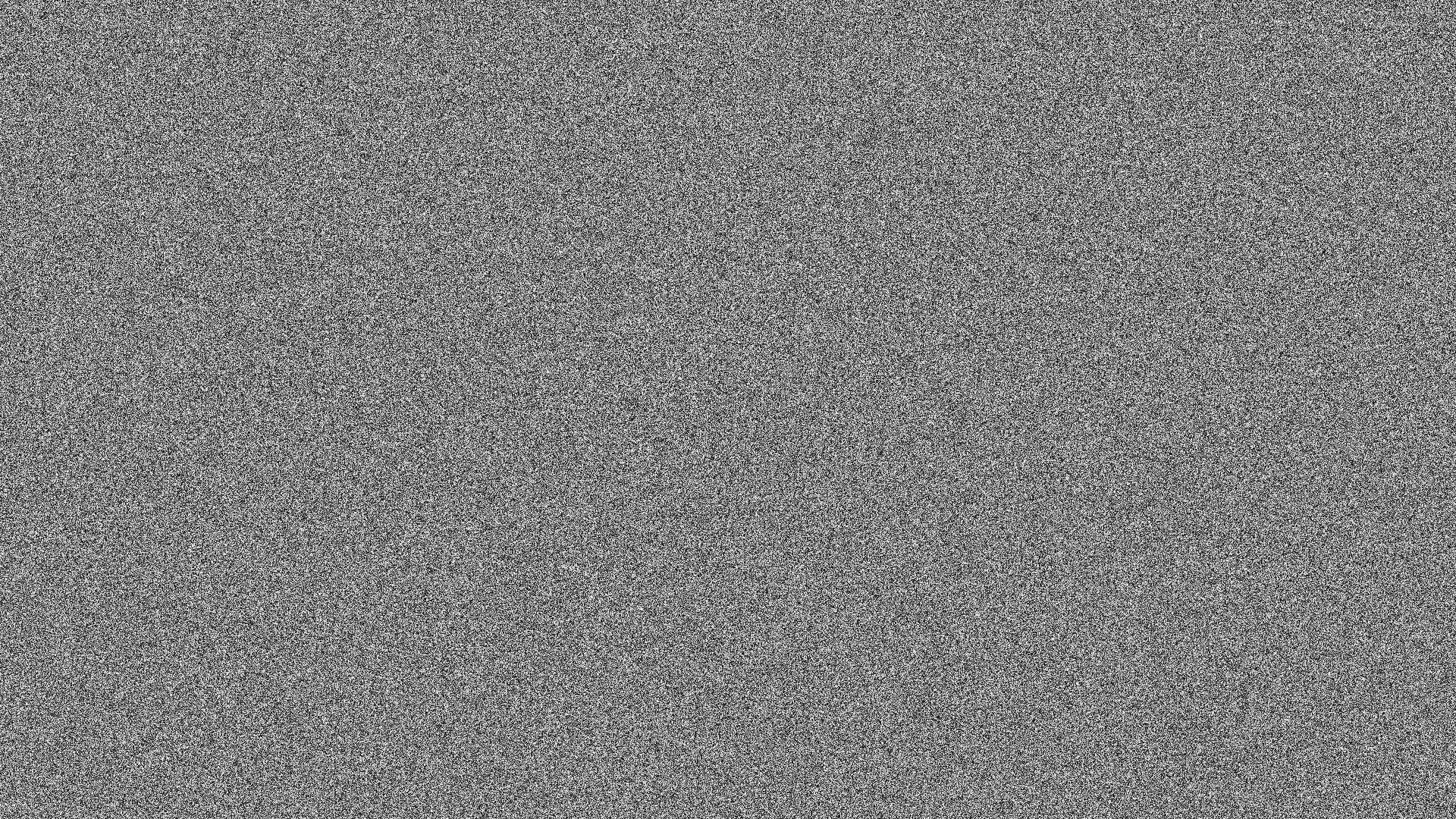}
		\caption{Initialization: noise}   
		\label{fig:init}
	\end{subfigure}
	\vskip\baselineskip
	\begin{subfigure}[b]{0.475\textwidth}   
		\centering 
		\includegraphics[width=\textwidth]{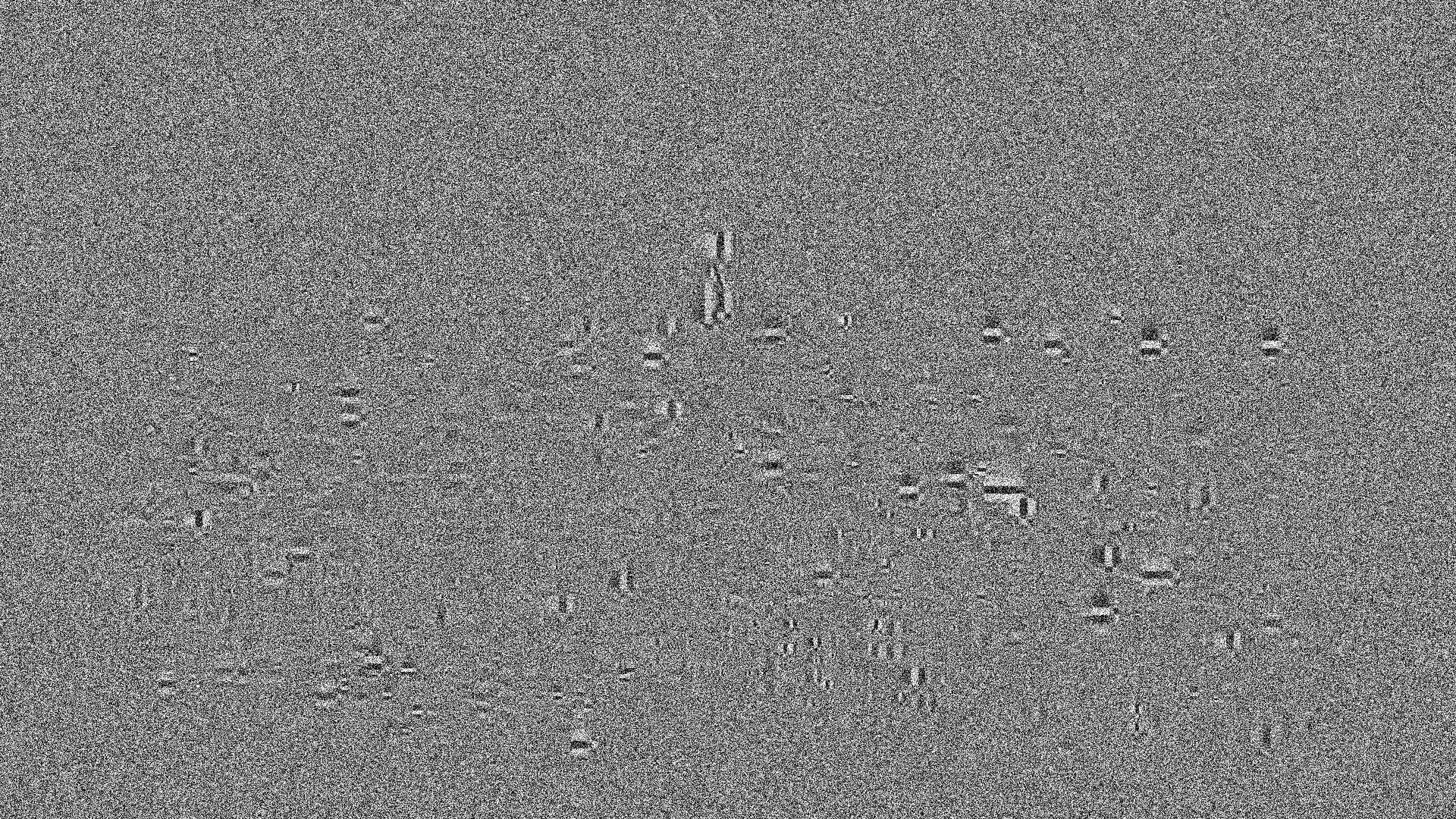}
		\caption{ADM optimized, metric value $1.000$.}    
		\label{fig:adm}
	\end{subfigure}
	\hfill
	\begin{subfigure}[b]{0.475\textwidth}   
		\centering 
		\includegraphics[width=\textwidth]{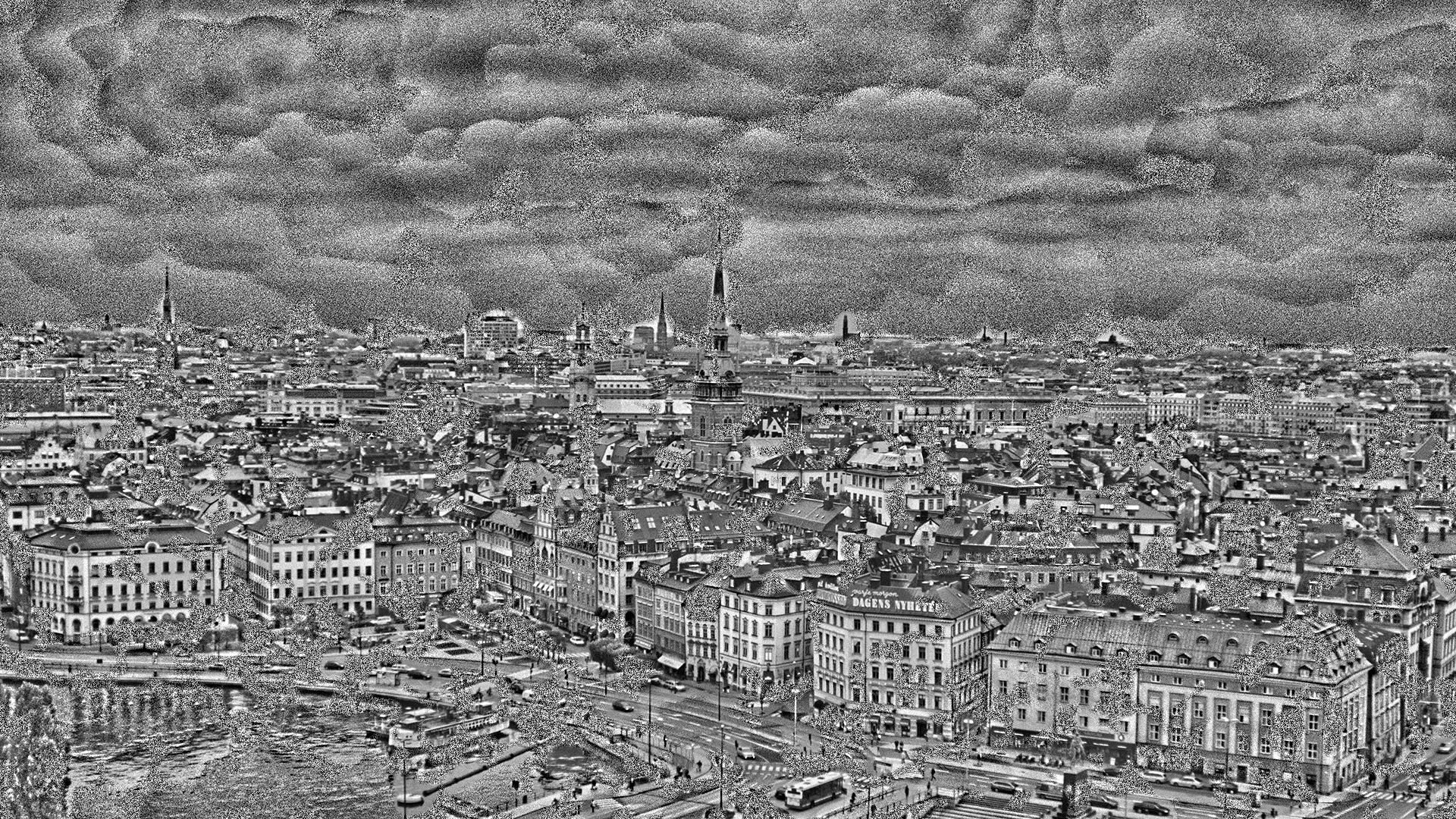}
		\caption{${\rm VIF}_0$ optimized, metric value $1.006$.}  
		\label{fig:vif0}
	\end{subfigure}
	\vskip\baselineskip
	\begin{subfigure}[b]{0.475\textwidth}
		\centering
		\includegraphics[width=\textwidth]{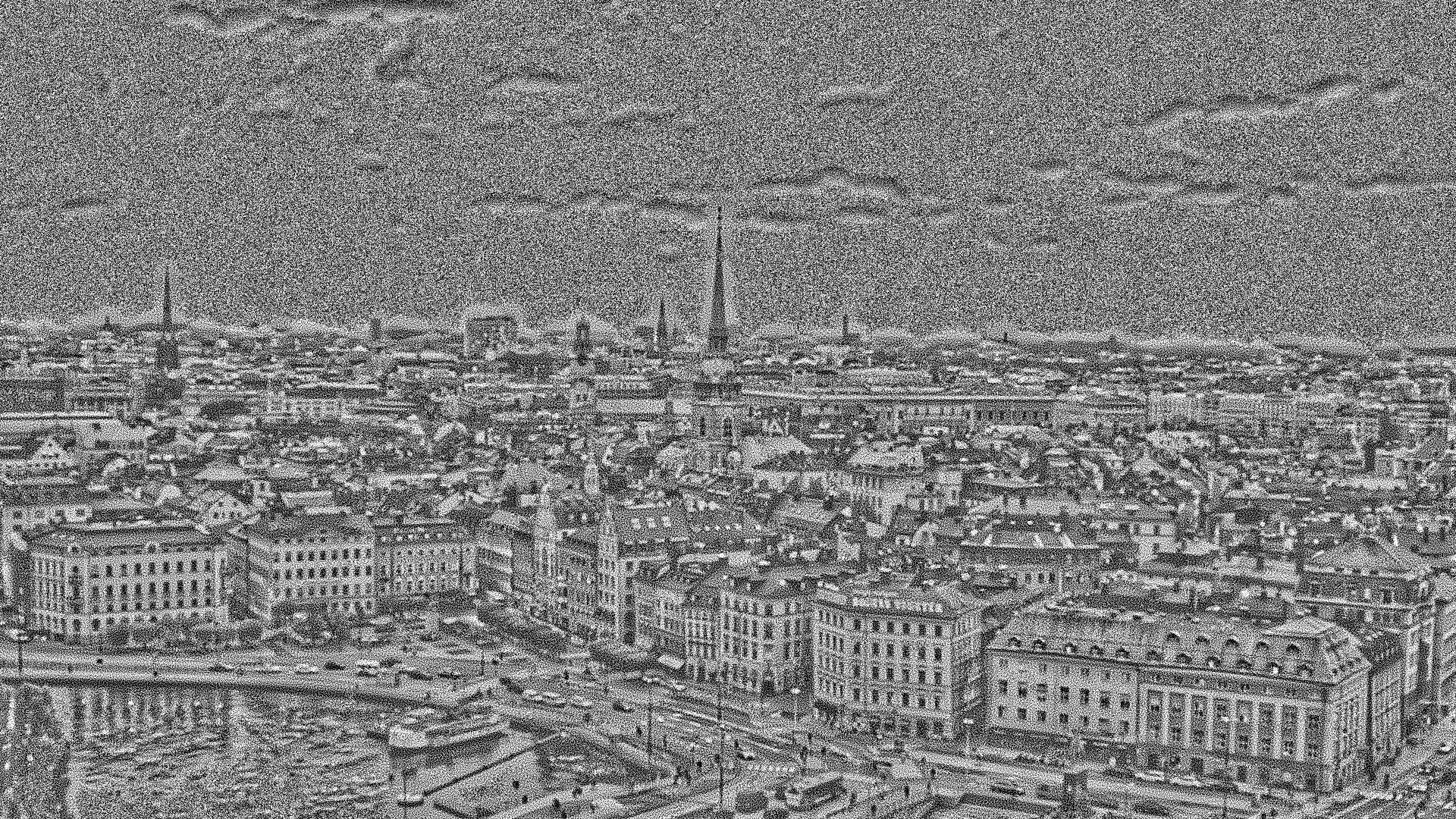}
		\caption{${\rm VIF}_{1}$ optimized, metric value $1.003$.}    
		\label{fig:vif1}
	\end{subfigure}
	\hfill
	\begin{subfigure}[b]{0.475\textwidth}  
		\centering 
		\includegraphics[width=\textwidth]{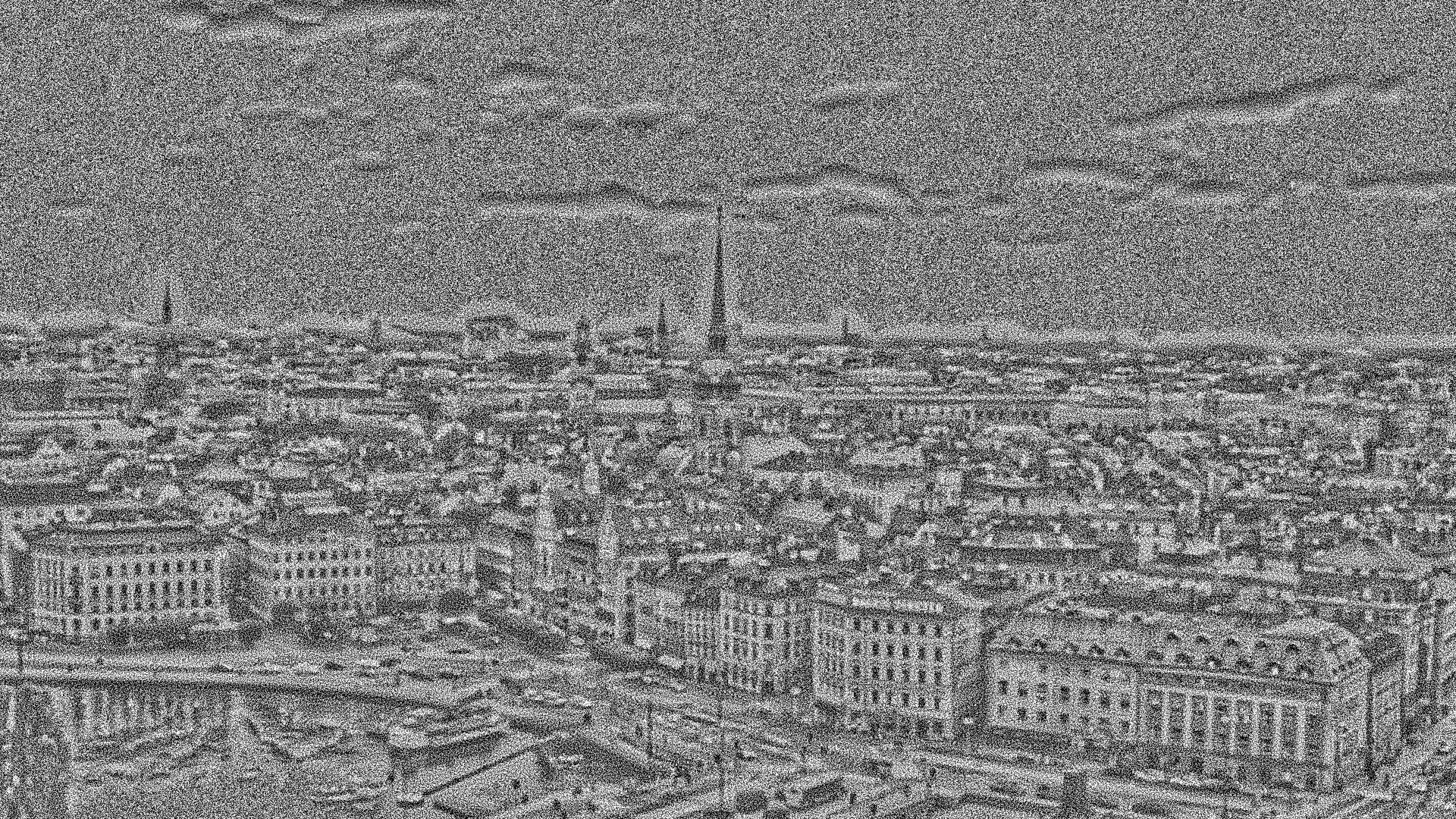}
		\caption{${\rm VIF}_{2}$ optimized, metric value $1.001$.}   
		\label{fig:vif2}
	\end{subfigure}
	\vskip\baselineskip
	\begin{subfigure}[b]{0.475\textwidth}   
		\centering 
		\includegraphics[width=\textwidth]{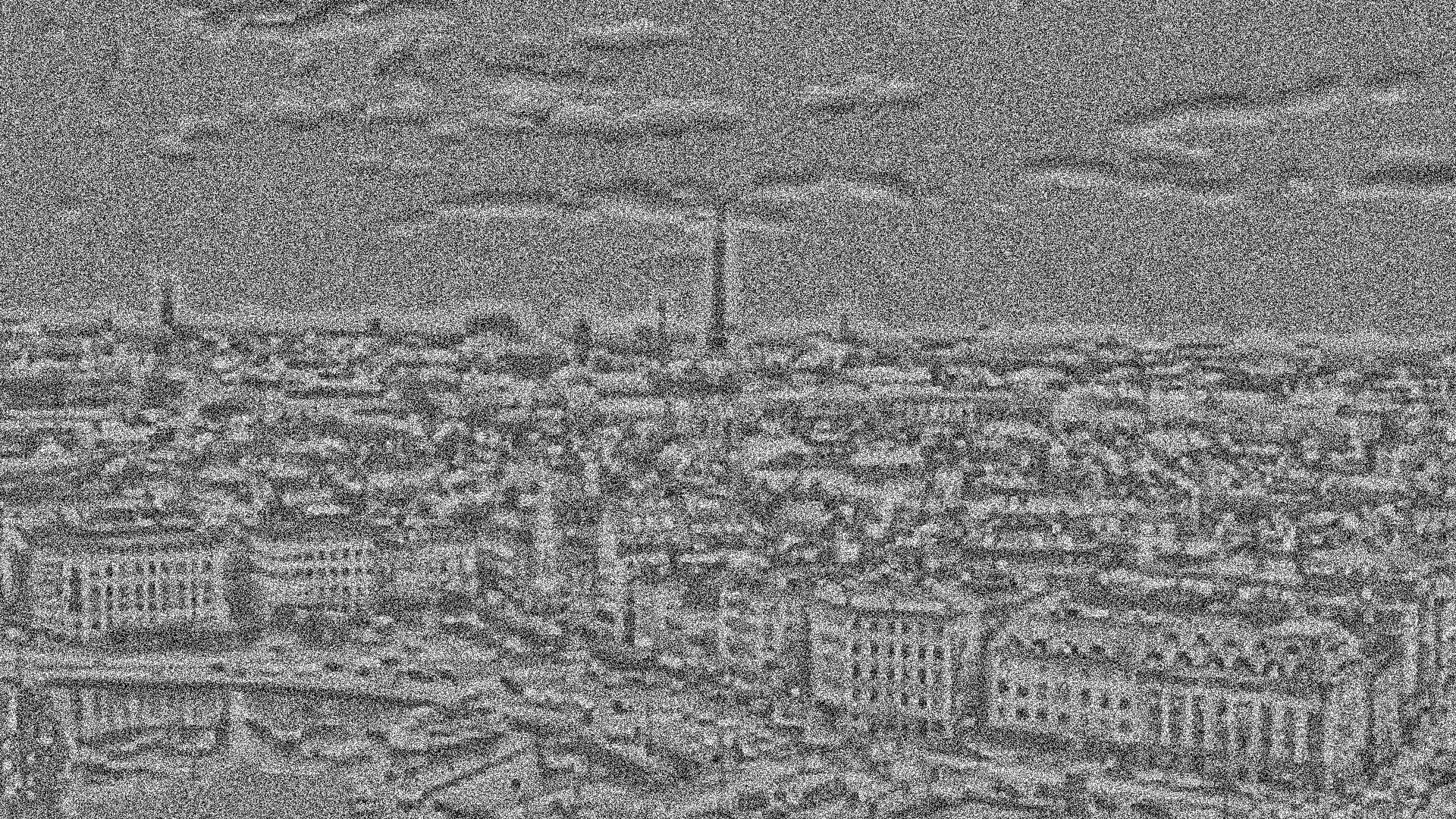}
		\caption{${\rm VIF}_{3}$ optimized, metric value $1.007$.}    
		\label{fig:vif3}
	\end{subfigure}
	\hfill
	\begin{subfigure}[b]{0.475\textwidth}   
		\centering 
		\includegraphics[width=\textwidth]{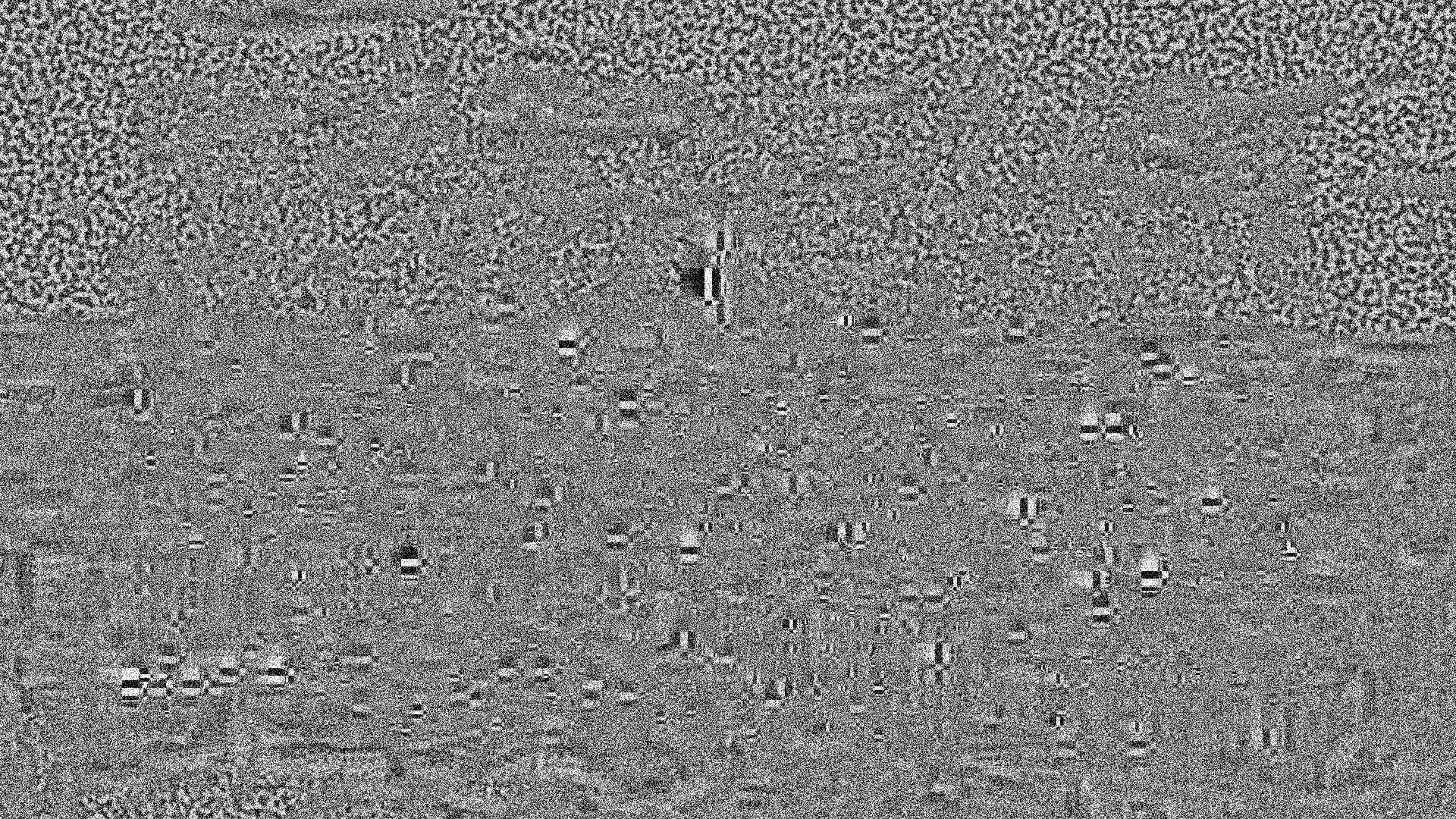}
		\caption{VMAF optimized, metric value $100.00$.}  
		\label{fig:vmaf}
	\end{subfigure}
	\caption{Reference image recovery: using the reference image (a) and choosing noise (b) as an initial condition we try to recover the reference image (a) by optimizing VMAF (h) or its elementary features (c)-(g) until they reach the values corresponding to the ideal quality; the final metric values are shown for each sub-metric in the captions.} 
	\label{fig:images_from_noise}
\end{figure*}

We apply this procedure for VMAF and its elementary submetrics: VIF for four scales (we denote them ${\rm VIF}_{0},
\ldots$, ${\rm VIF}_{3}$ and ADM. $I_0$ is initialized with uniform noise on $[0,255]$ or, alternatively, with a
compressed version of the image; only the Y component of the image is used. We employ the Adam optimizer with 
learning rate from $0.1$ to $1$ depending on the metric and hyperparameters $\beta_{1}=0.9$ and $\beta_{2}=0.999$.
The optimization is stopped when the target metric reaches $100$ for VMAF or $1$ for VIF and ADM, as our goal is to recover the original image without additional enhancement. The results of the restoration for each submetric are presented in Fig. \ref{fig:images_from_noise}. It is seen that ADM was unsuccessful in restoring the images, while VIF demonstrated some resemblance of the restored and the reference image depending on the scale. These results are in correspondence to those of \cite{Ding_2021}, where the restoration using VIF was more or less successful\footnote{Problems appeared in the regions related to the sky; see \cite{Ding_2021}, pp. 3-4.}. Overall it should be admitted that restoration using the full VMAF metric is not satisfactory. Some additional information can be obtained when portreying spectra for the corresponding images, they are shown in Fig. \ref{fig:restoration_spectra}.
It is clearly seen the resemblance of  the spectra for the original image and the one resotred using VIF quality metric. Unlike them the spectrum of the image restored using VMAF differs significantly from the other two, demonstrating abnormal growth in mid-frequency as well as in high frequency ranges. These results are in correspondence with Fig. \ref{fig:images_from_noise}. Unsatisfactory restoration is partially accounted for by the results for the ADM metric: while VIF produces the restored image close to the original, ADM fails in restoration which, in turn, results in the bad restoration for VMAF. The results suggest that VIF turns out to be more stable from the point of view of the learning algorithm. It also seems not so easy to comare VIF to ADM as the former was implemented in pixel space both in its original version and in PyTorch version \cite{vmaf_torch}. On the other hand, these results enable to see significant discrepancies between the subjective quality and the objective metric scores: obviously the restored images in Fig. \ref{fig:images_from_noise} have poor quality in disagreement to the score.

\begin{figure*}
	\centering
	\begin{subfigure}[b]{0.475\textwidth}
		\centering
		\includegraphics[width=\textwidth]{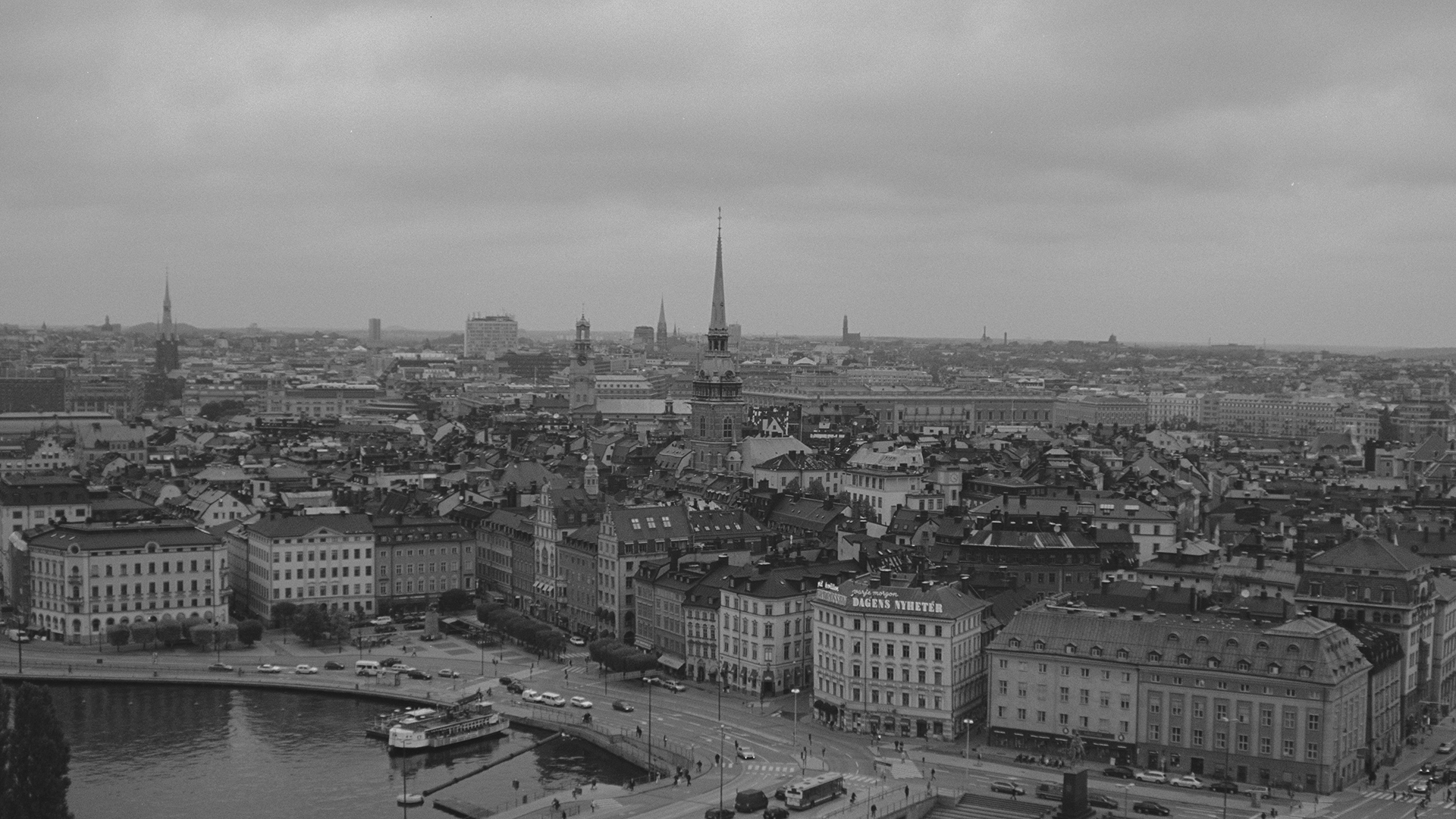}
		\caption{Reference image}    
	\end{subfigure}
	\hfill
	\begin{subfigure}[b]{0.475\textwidth}  
		\centering 
		\includegraphics[width=\textwidth]{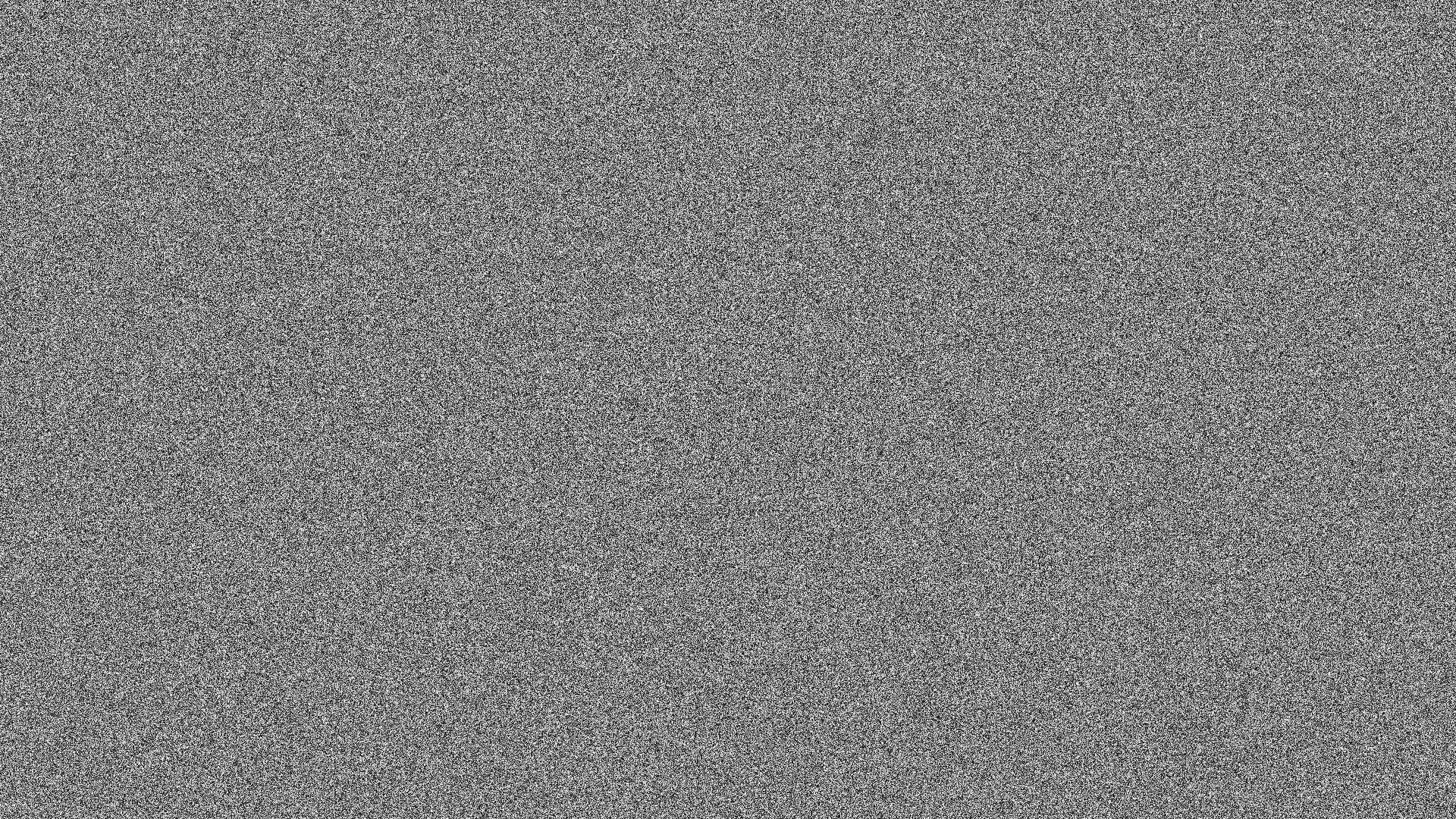}
		\caption{Initialization: noise}   
	\end{subfigure}
	\vskip\baselineskip
	\begin{subfigure}[b]{0.475\textwidth}   
		\centering 
		\includegraphics[width=\textwidth]{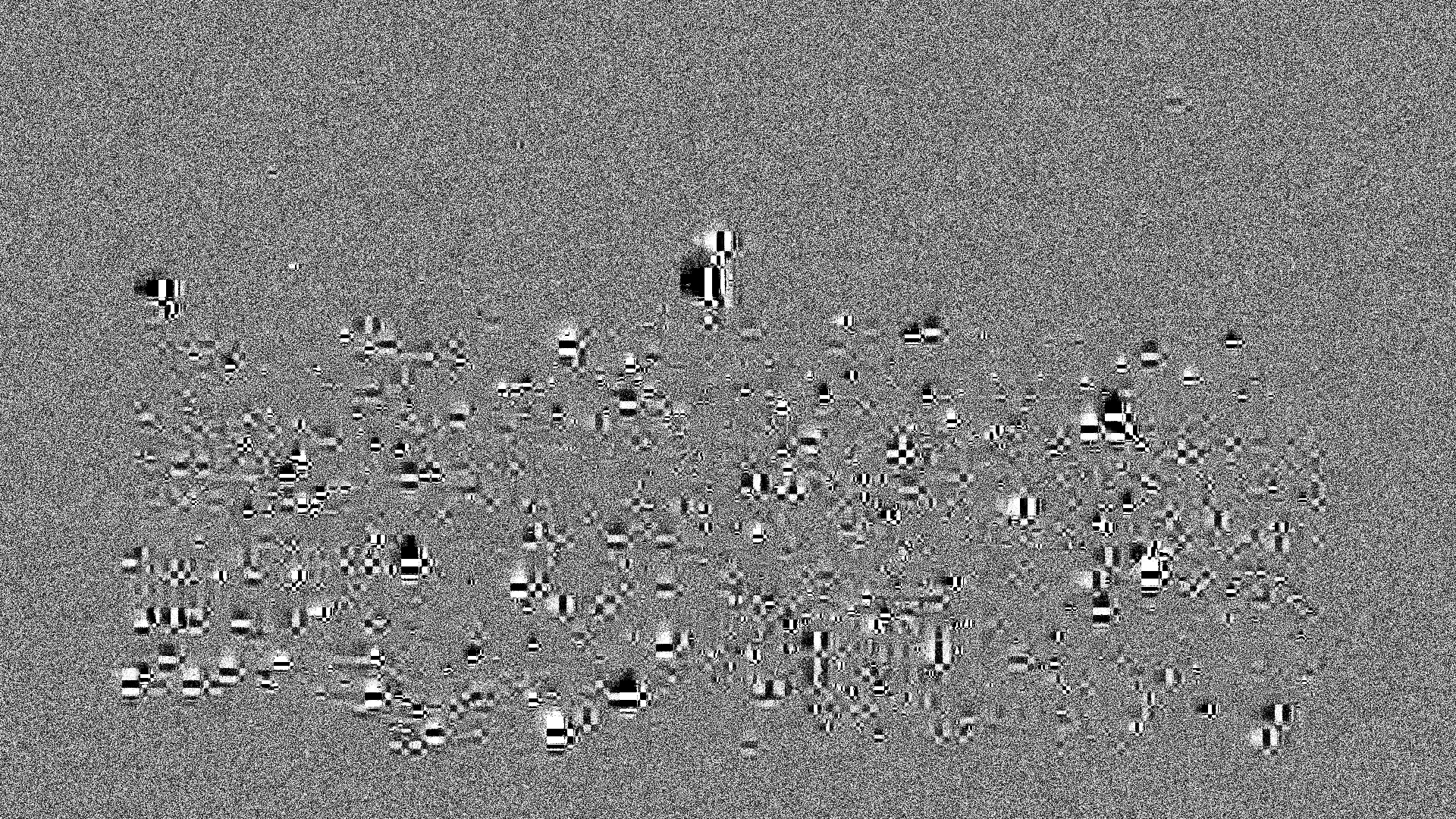}
		\caption{ADM optimized, metric value $1.626$.}    
	\end{subfigure}
	\hfill
	\begin{subfigure}[b]{0.475\textwidth}   
		\centering 
		\includegraphics[width=\textwidth]{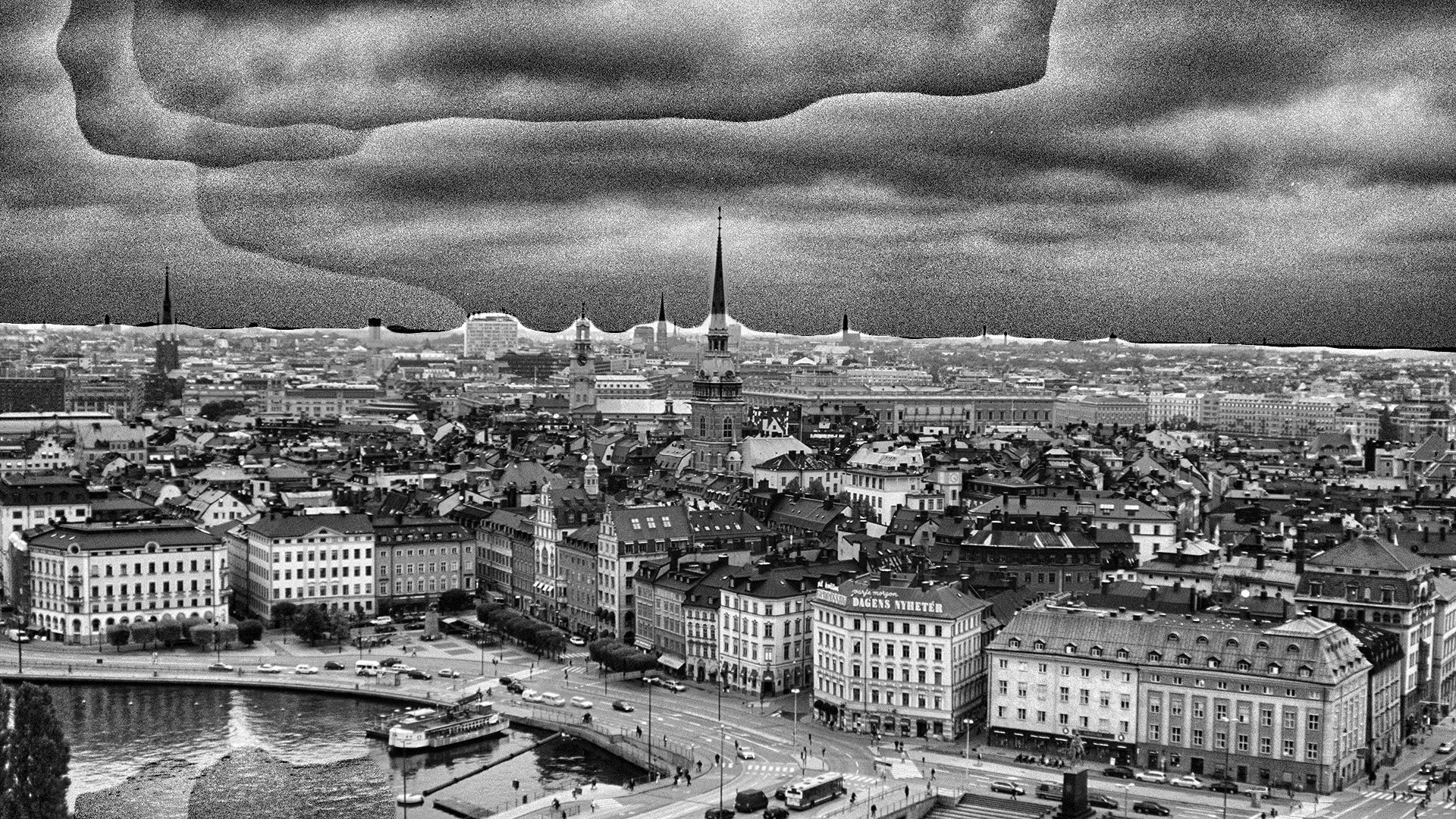}
		\caption{${\rm VIF}_0$ optimized, metric value $1.659$.}  
	\end{subfigure}
	\vskip\baselineskip
	\begin{subfigure}[b]{0.475\textwidth}
		\centering
		\includegraphics[width=\textwidth]{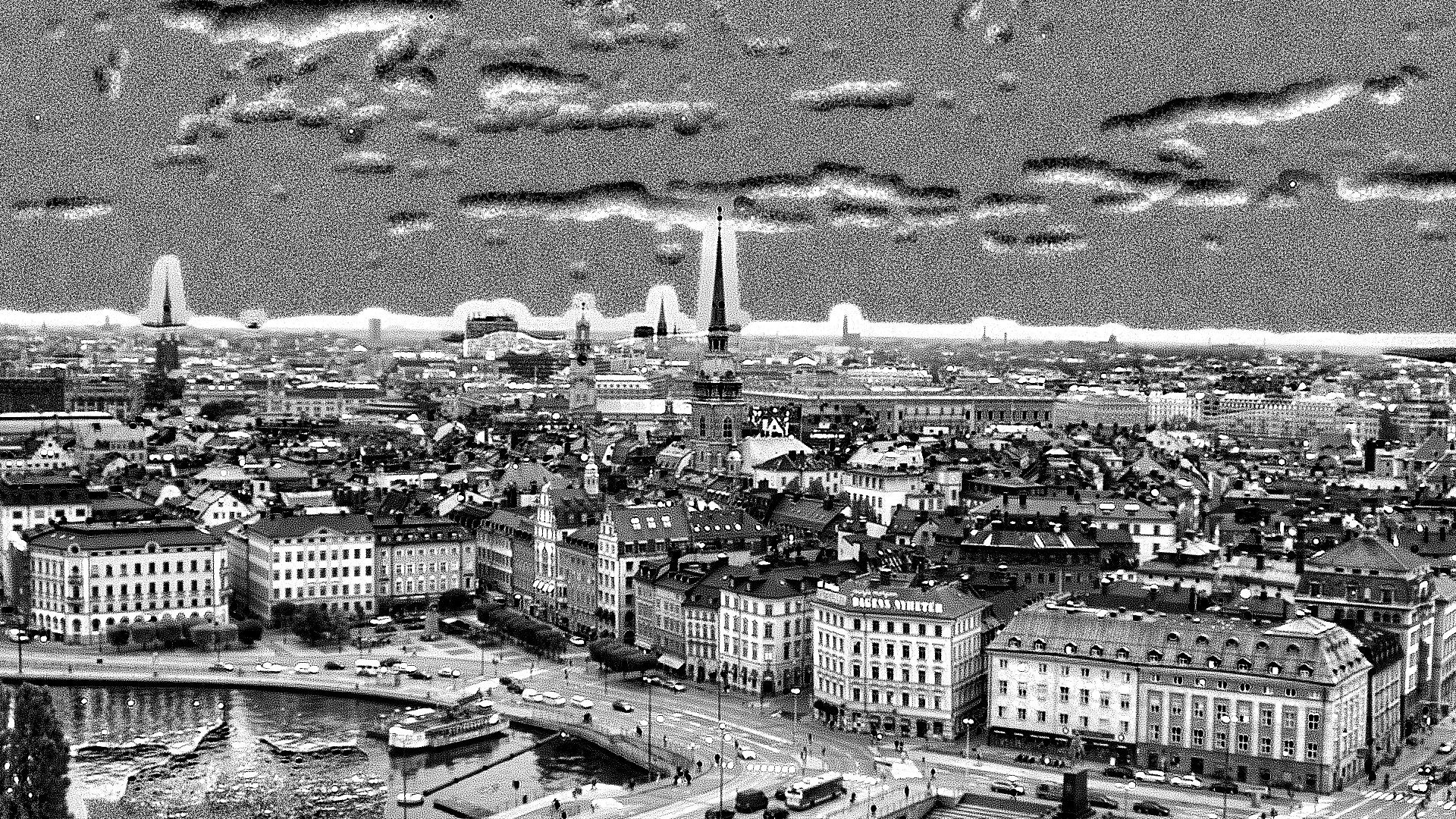}
		\caption{${\rm VIF}_{1}$ optimized, metric value $1.501$.}    
	\end{subfigure}
	\hfill
	\begin{subfigure}[b]{0.475\textwidth}  
		\centering 
		\includegraphics[width=\textwidth]{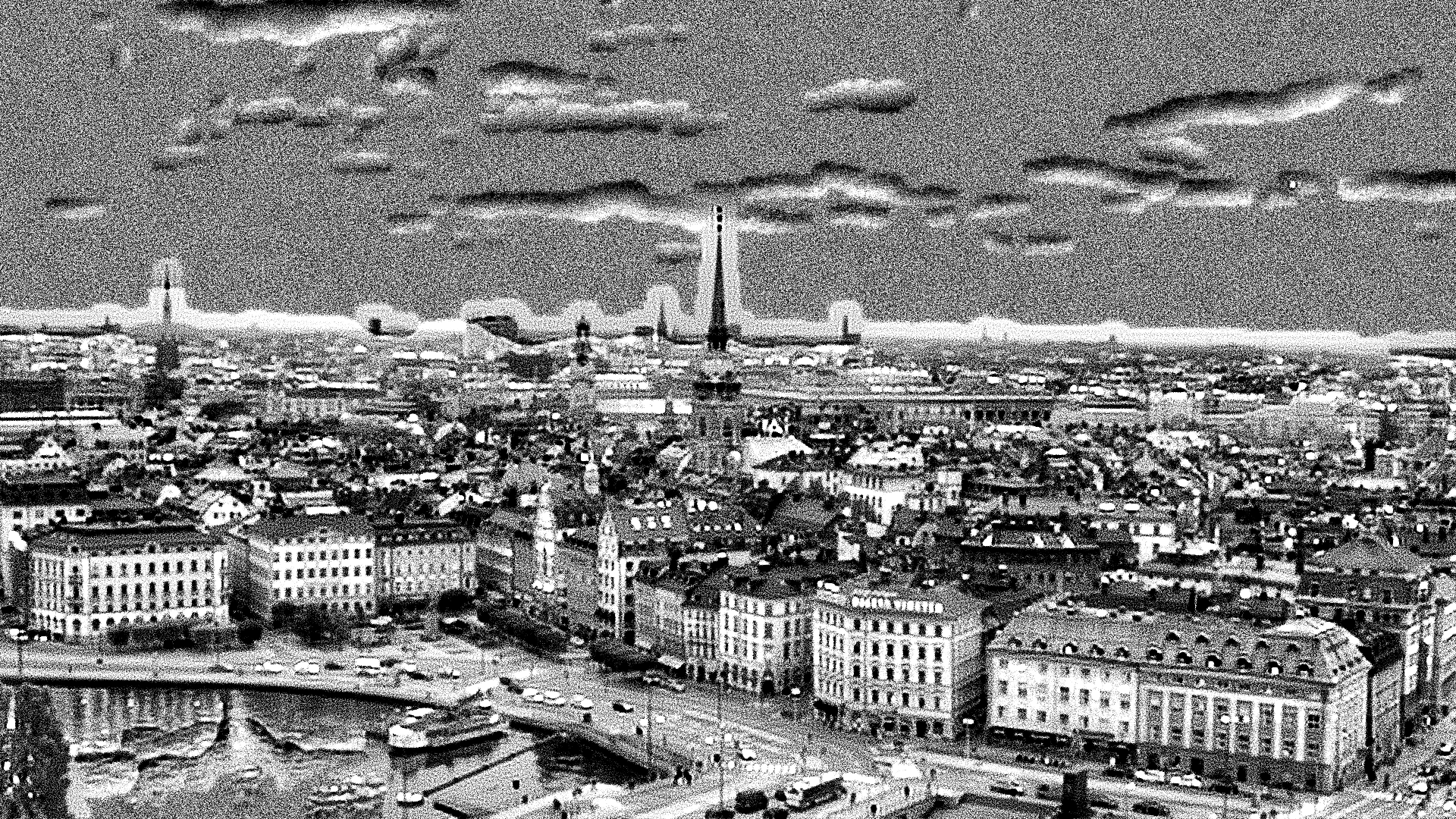}
		\caption{${\rm VIF}_{2}$ optimized, metric value $1.550$.}   
	\end{subfigure}
	\vskip\baselineskip
	\begin{subfigure}[b]{0.475\textwidth}   
		\centering 
		\includegraphics[width=\textwidth]{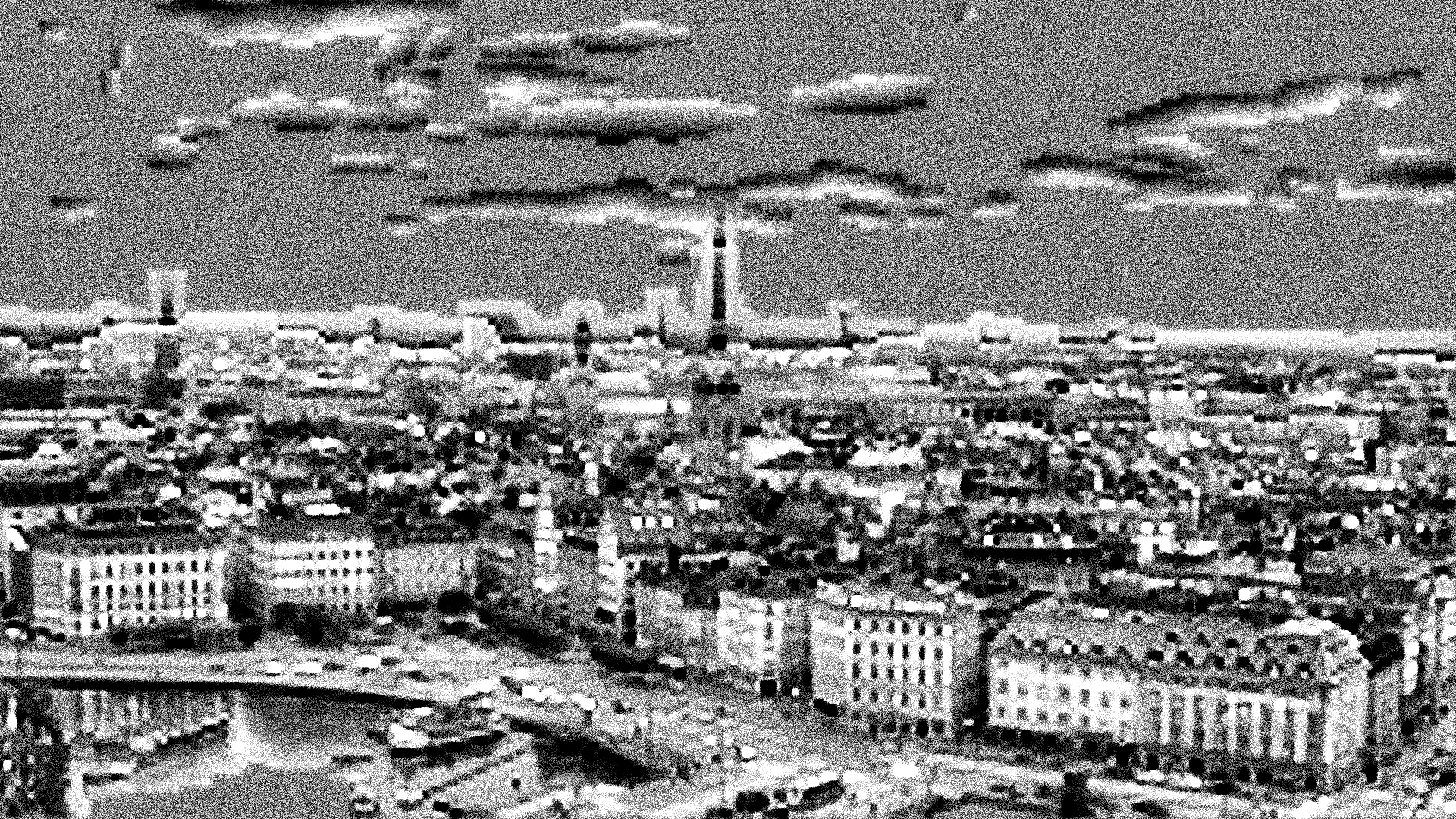}
		\caption{${\rm VIF}_{3}$ optimized, metric value $1.632$.}    
	\end{subfigure}
	\hfill
	\begin{subfigure}[b]{0.475\textwidth}   
		\centering 
		\includegraphics[width=\textwidth]{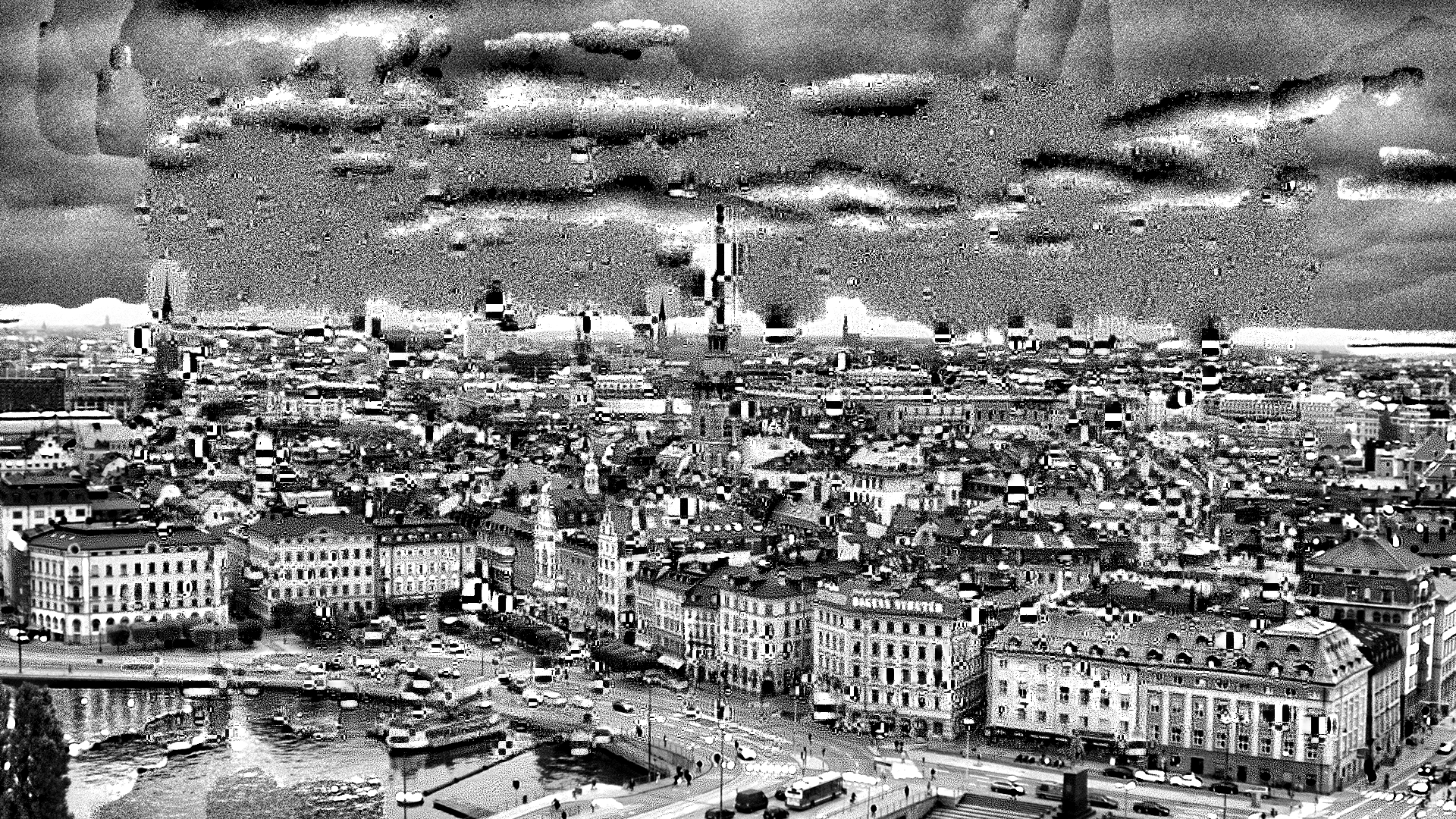}
		\caption{VMAF optimized, metric value $304.93$.}  
	\end{subfigure}
	\caption{Reference image recovery: the set-up of the experiment is the same as shown in Fig. \ref{fig:images_from_noise} except the learning process was not stopped when the metrics attained their limiting values ($100$ for VMAF and $1$ for VIF and ADM). The limiting values of the metrics are shown in the captions for the subfigures.} 
	\label{fig:images_from_noise_unclipped}
\end{figure*}

We also conducted the same experiment without stopping at the threshold $100$ for VMAF or $1$ for VIF and ADM. Instead, the optimization continued until the target metric stops its growth and convergence is reached. The results are presented in Fig. \ref{fig:images_from_noise_unclipped}. Comparing the results for VMAF to Fig. \ref{fig:images_from_noise} it is seen that the restoration was noticeably improved wrt the previous experiement despite the visual quality of ADM enhanced image remains extremely poor.  These results are also confirmed by the corresponding power spectra shown in Fig. \ref{fig:restoration_spectra} which demonstrate much more similarity to the original spectrum in all frequency scales compared to the case when restoration stopped at $100$ for VMAF and $1$ for VIF. However the discrepancies compared to the reference image observed even in this case  presumably reflect the quality of models used to construct VIF and ADM metrics. It is instructive to look at the image recovered with $100$ VMAF units (Fig.  \ref{fig:images_from_noise}): its quality is drastically poor. This indicates that the perturbation obtained through the optimization procedure is favorable to VMAF but corresponds to the regime in which VMAF model becomes invalid.

It is interesting to note that even when attained the maximum value for submetric the quality of the restored images is not satisfactory. This is especially noticeable for ${\rm VIF}_{0}$ where some regions in the figure were obviously not recovered and remained unstructured noisy regions which can be observed by more close inspection of the Fig. \ref{fig:images_from_noise}d. Continuing the process in the regime of enhancement (${\rm VIF}_{0} > 1$) we were able to remove the obvious noisy regions for ${\rm VIF}_{0}$  but this required significant metric increase up to $\approx 1.6$. It is also noticeable that when increasing the scale in VIF the images become blurrier. This is accounted for by the presense of the gaussian filtering at various scales in VIF as well as by downsampling.

ADM in its turn is able to recover almost no features of the original image, while manifesting a different type of artifacts. 
For ADM it seems plausible to assume that these artifacts arise as a result of wavelet transform as they resemble the wavelet basis functions\footnote{See, e.g., https://cseweb.ucsd.edu/classes/fa20/cse166-a/lec12.pdf}.
Images recovered using VMAF as the target function show a mix of both types of artifacts.

It is known that VMAF contains several sub-metrics, some of them were used above. Another sub-metric is related to temporal motion. Whenever we apply VMAF to videos we do computations with the full set of submetrics, but if we compute VMAF for images we simply set the motion feature to zero.
This is equivalent to treating images as videos with a single frame, since the motion feature for the first frame in any video is normally set to zero. 
Note that for the case of motion feature equal to zero, i.e., for single images or videos with still frames with no temporal changes, the VMAF score for two identical images is not equal to $100$ but approximately $97.4$.  
In our experiments with images we set the motion feature to zero. For videos identical results are expected since the motion feature is computed using only the reference image, so the gradients with respect to the distorted image are not affected by it.

\begin{figure*}
	\centering
	\begin{subfigure}[b]{0.475\textwidth}
		\centering
		\includegraphics[width=\textwidth]{figures/reference.png}
		\caption{Reference image}    
	\end{subfigure}
	\hfill
	\begin{subfigure}[b]{0.475\textwidth}  
		\centering 
		\includegraphics[width=\textwidth]{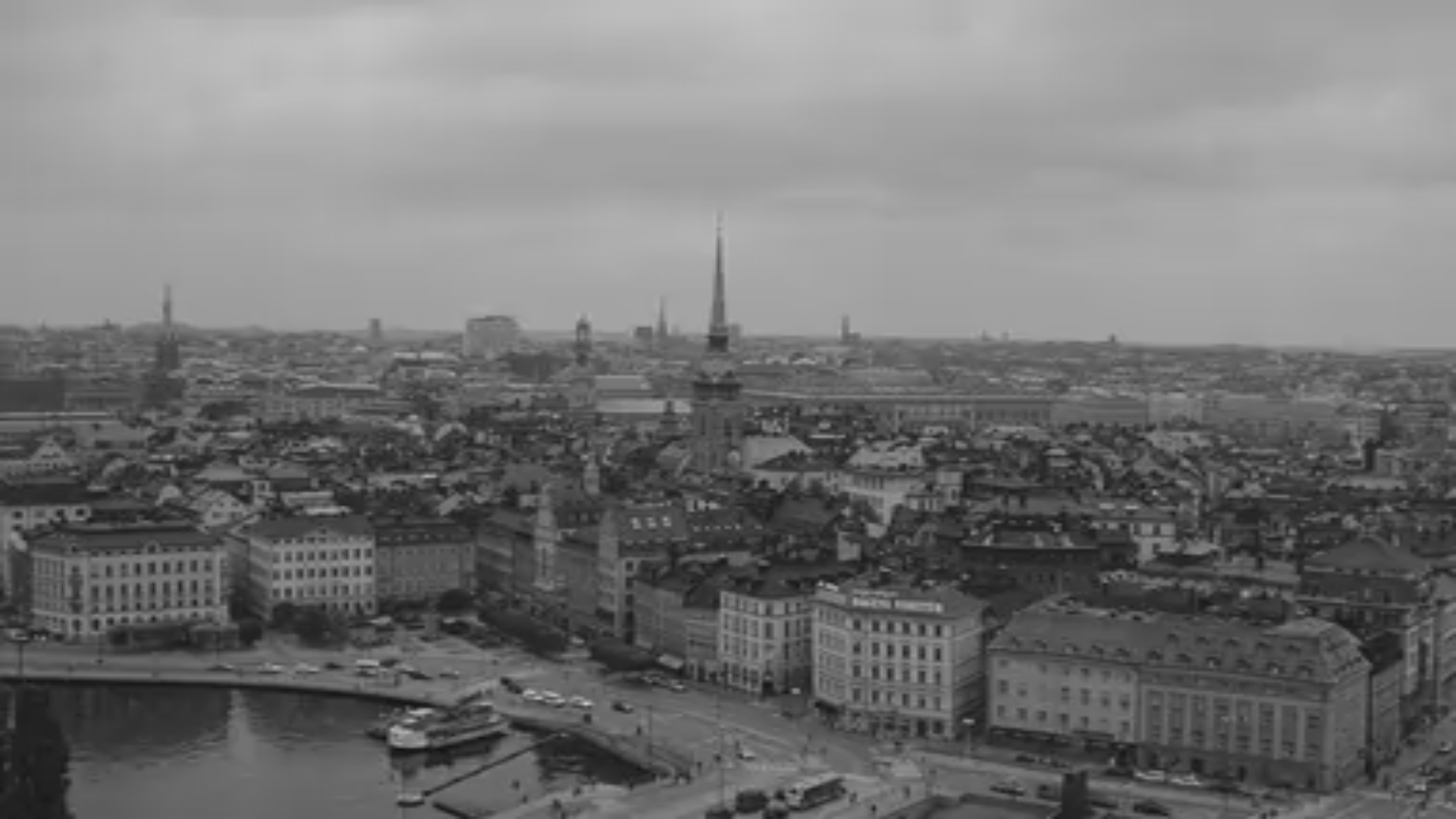}
		\caption{Initialization: compressed version of the reference image}   
	\end{subfigure}
	\vskip\baselineskip
	\begin{subfigure}[b]{0.475\textwidth}   
		\centering 
		\includegraphics[width=\textwidth]{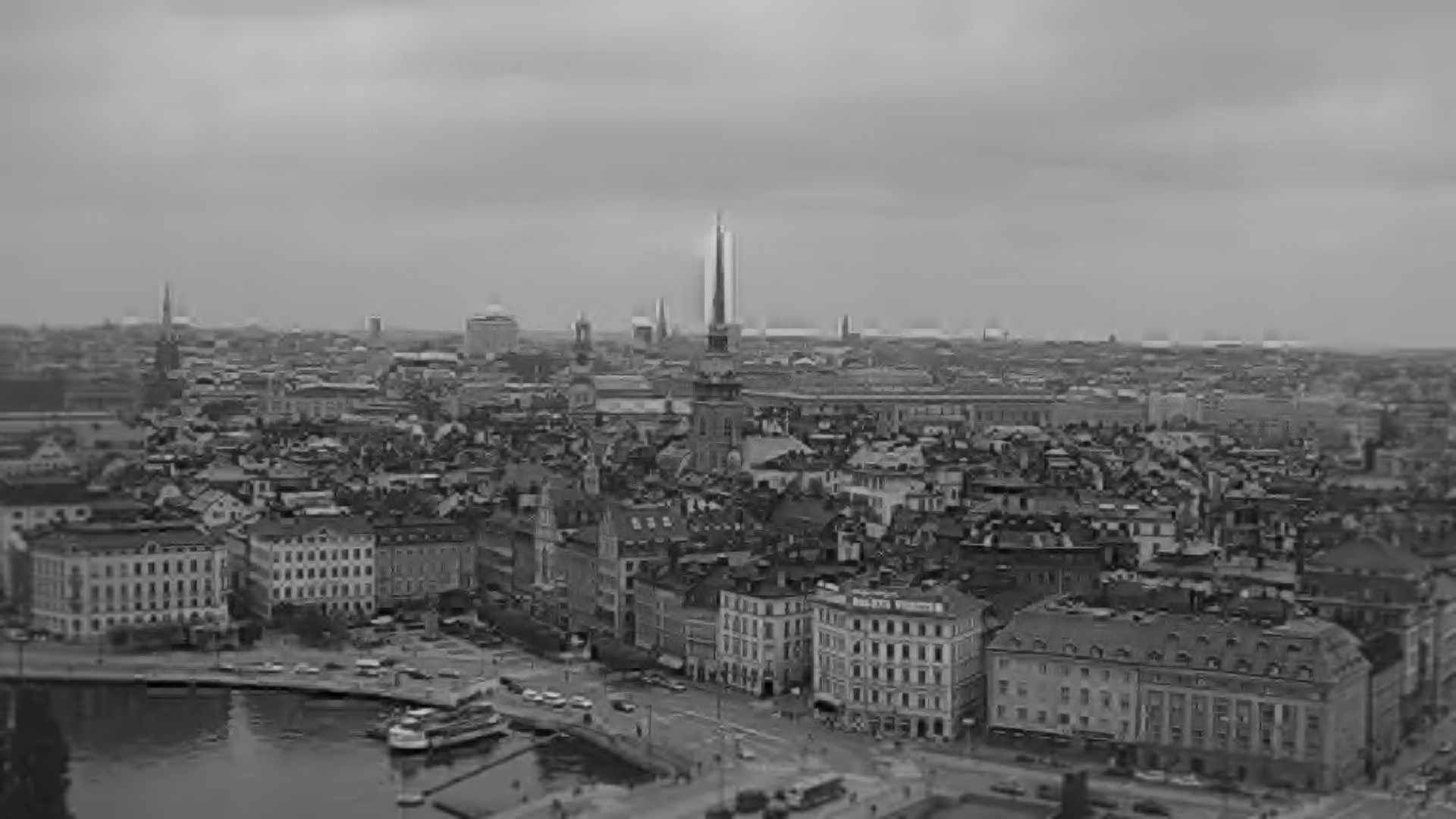}
		\caption{ADM optimized, metric value $1.010$.}    
	\end{subfigure}
	\hfill
	\begin{subfigure}[b]{0.475\textwidth}   
		\centering 
		\includegraphics[width=\textwidth]{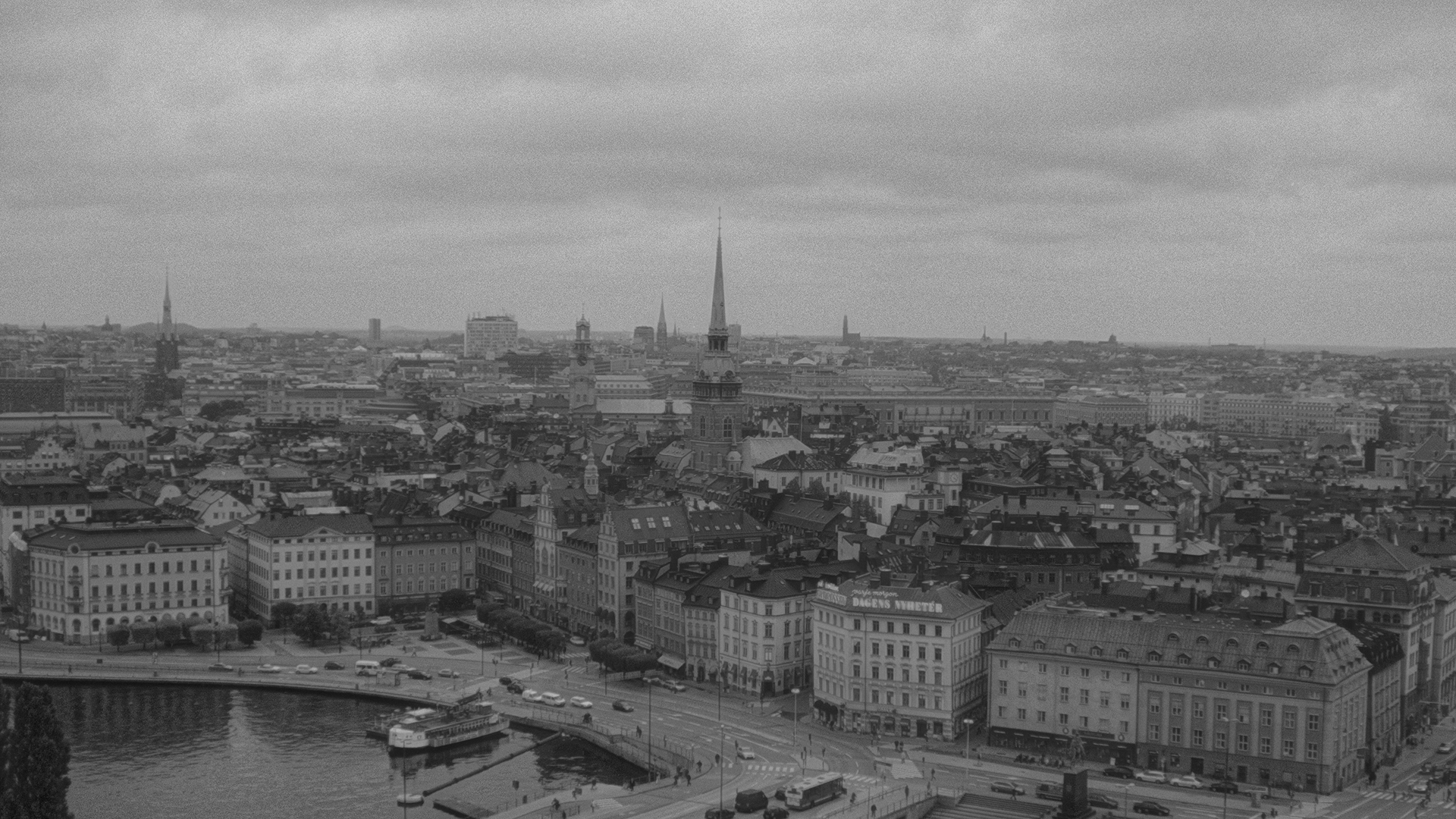}
		\caption{${\rm VIF}_0$ optimized, metric value $1.003$.}  
	\end{subfigure}
	\vskip\baselineskip
	\begin{subfigure}[b]{0.475\textwidth}
		\centering
		\includegraphics[width=\textwidth]{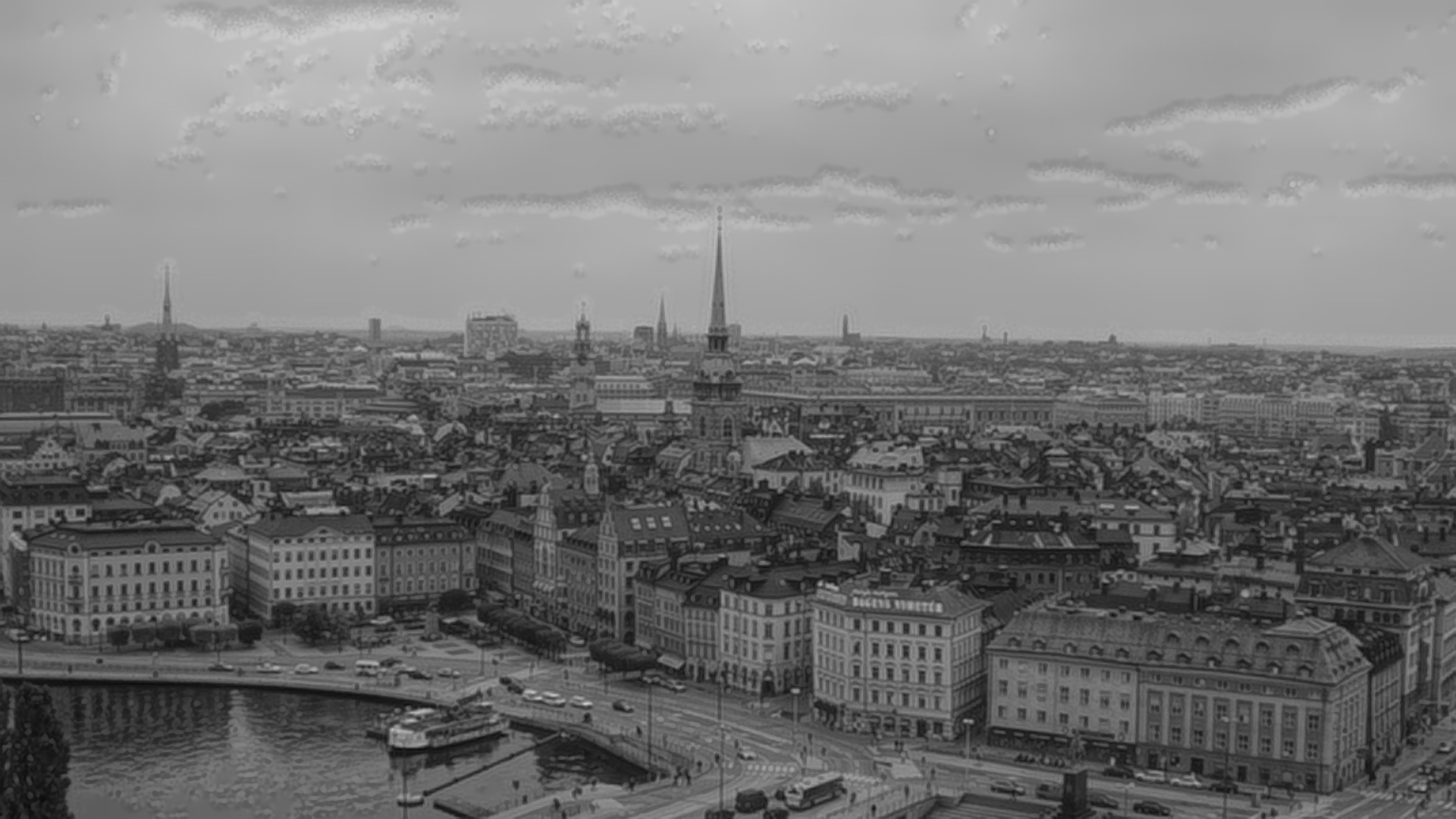}
		\caption{${\rm VIF}_{1}$ optimized, metric value $1.002$.}    
	\end{subfigure}
	\hfill
	\begin{subfigure}[b]{0.475\textwidth}  
		\centering 
		\includegraphics[width=\textwidth]{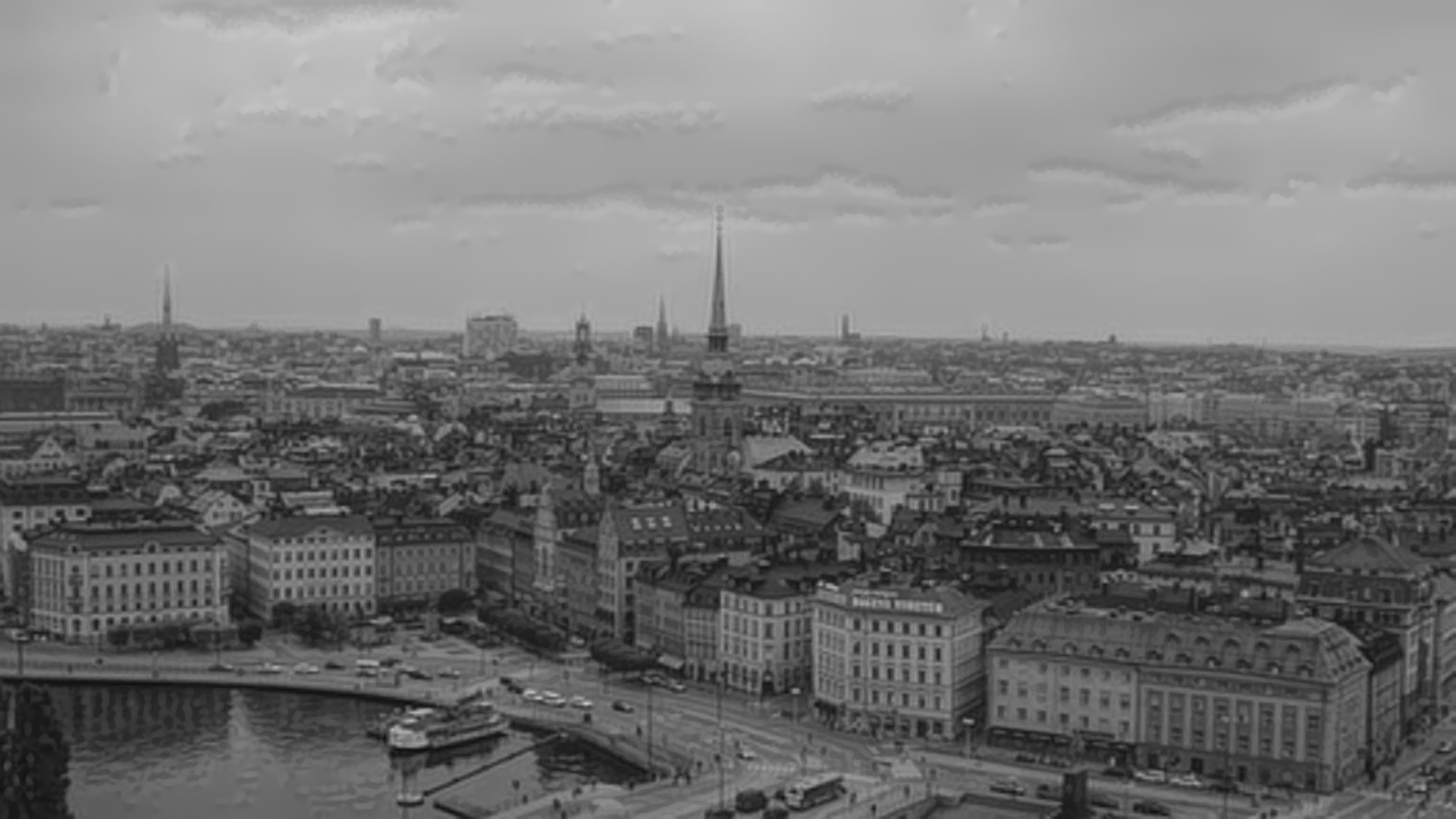}
		\caption{${\rm VIF}_{2}$ optimized, metric value $1.002$.}   
	\end{subfigure}
	\vskip\baselineskip
	\begin{subfigure}[b]{0.475\textwidth}   
		\centering 
		\includegraphics[width=\textwidth]{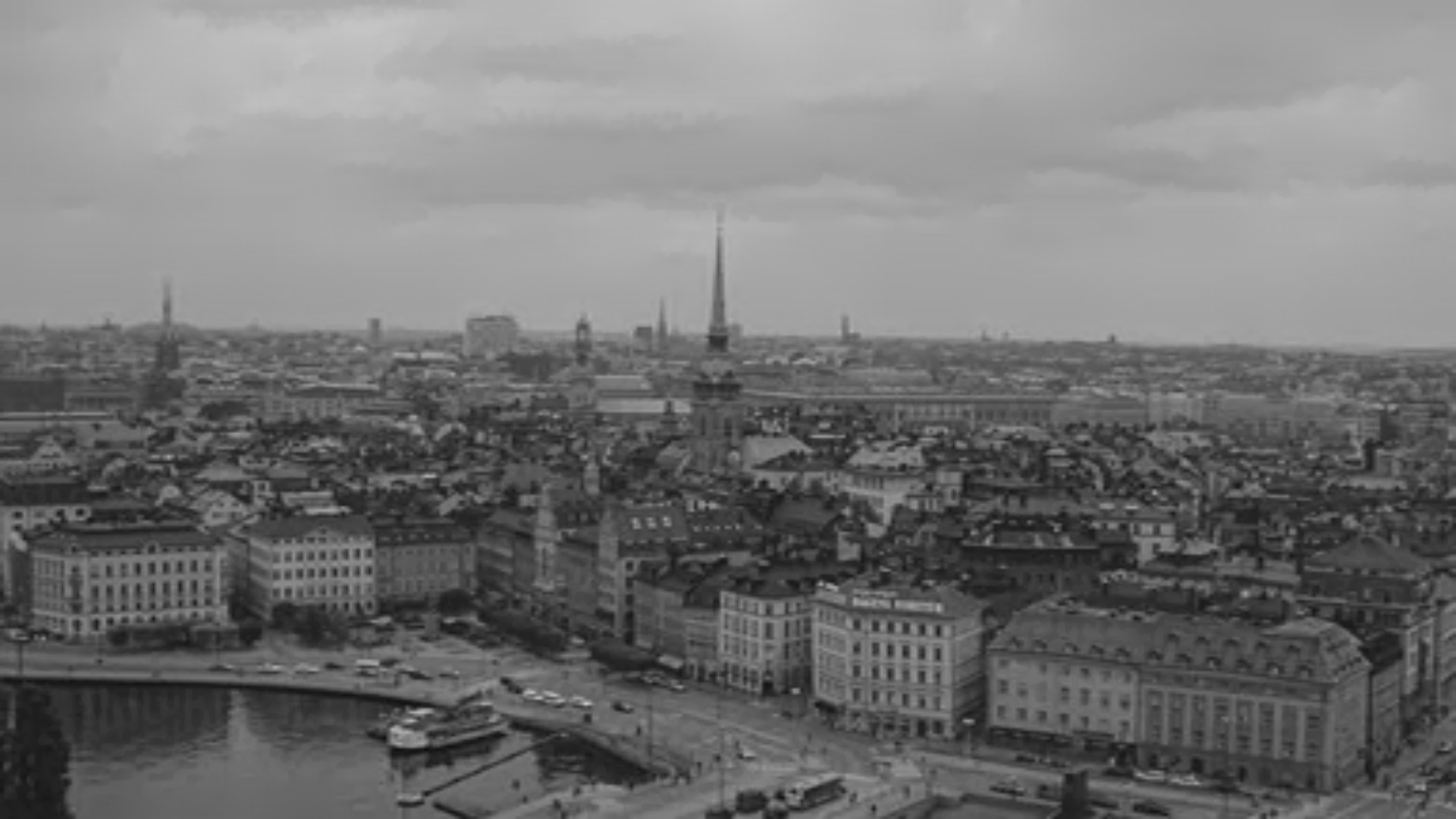}
		\caption{${\rm VIF}_{3}$ optimized, metric value $1.006$.}    
	\end{subfigure}
	\hfill
	\begin{subfigure}[b]{0.475\textwidth}   
		\centering 
		\includegraphics[width=\textwidth]{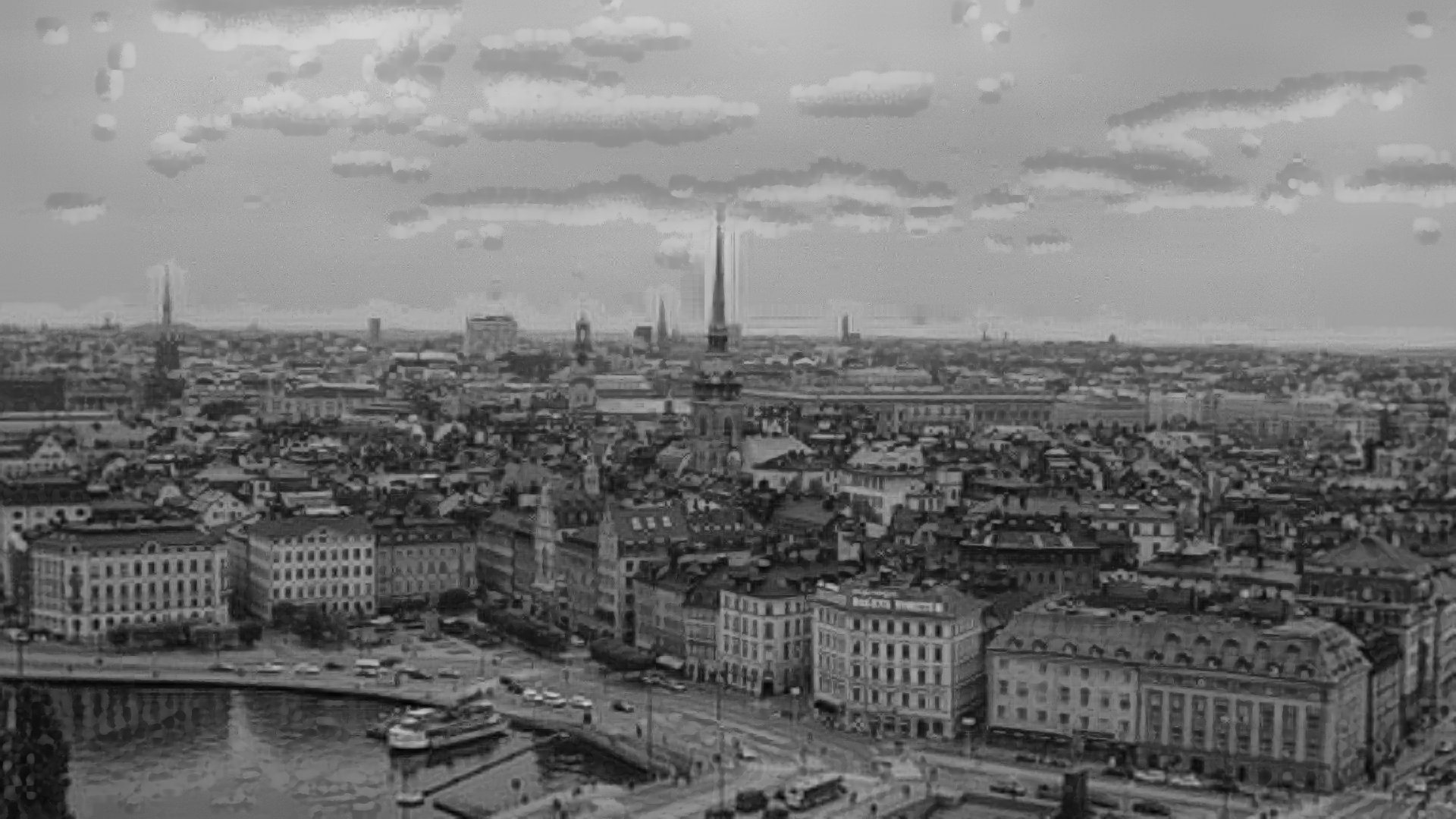}
		\caption{VMAF optimized, metric value $100.00$.}  
	\end{subfigure}
	\caption{Reference image recovery: starting from the compressed version of the reference image (b) we recover the reference image (a) by optimizing VMAF (h) or its elementary features (c)-(g), until they reach values indicating "perfect" quality, the final metric values are shown for each sub-metric.} 
	\label{fig:images_from_compressed}
\end{figure*}

For the case of initial image being the compressed version of the reference one, the results are shown in Fig. \ref{fig:images_from_compressed}. Now VIF at scale $0$ is actually able to restore the details of the image lost due to compression. However, some high-frequency noise is added to the image; this can be caused by the fact that image is smoothed by a gaussian filter, so that noisy and non-noisy version of the image are indistinguishable to VIF. VMAF, ADM and VIF for other scales are not able to recover the image and also add additional artifacts.

This experiment presents a case where the score of VMAF metric significantly differs from human judgement: for the image shown in Fig. \ref{fig:vmaf} visual quality is noticeably poor, yet the VMAF score is $100$.

This experiment also provides some evidence that using VMAF as a loss (without some other additional losses) in image/video enhancement tasks (e.g. super-resolution, denoising, deblurring) may not be optimal since the target network may learn to produce the types of  artifacts described above.

\section*{Discussion}

\subsection*{Concerning the perturbation models}

The results of this work can also be viewed in a different perspective. When dealing with adversarial perturbations it is noticed that they are many in number and extremely different in their properties. This raises the question of classification or modelling of the perturbations. The ideas on creating perturbation models for some types of neural networks were investigated in \cite{madry2019deep}. Our analysis of adversarial perturbations aims at contributing to the modelling of the latter for images and related image quality metrics. The analysis we presented on finding the properties of perturbations which can maximize VMAF or its submetrics allows to describe the characteristic properties of the perturbations which were shown to be invariant on the perturbation amplitudes. We think of this as one preliminary step in the problem of perturbation modelling and hope to elaborate these models in the subsequent papers. On the other hand, we believe that this approach goes to the line of research proposed in a number of recent works (see, e.g.,  \cite{madry2019deep}). It is necessary to stress, however, that if for the large-scale networks, for which the problem of vulnerability due  to adversarial perturbations originally emerged, the approach concerned with the direct investigation of the properties of the pertubrations may be insufficient and hence the saddle-point approach is fully justified, for more restricted problems like image quality and relatively simple losses the analysis of perturbations properties can provide additional insights and thus is thought to be useful.

\subsection*{Some remarks concerning abnormal metric behavior}

One of the natural questions which can be raised in connection with this work is the problem of so called metric breaking. Simplistically speaking, it poses the following problem. If we have a processing method which alters the image, and if we use VMAF or any other quality measure as our main means to judge about the relative quality of the image before and after the alteration, it is claimed that one can ``deceive'' the metric trying to improve VMAF score by applying various techniques, e.g., those described above. Putting aside the jargon (metric breaking, deceive etc.), such problem formulation seems to be too narrow and unilateral. Objectively, we rely our judgements about the image quality on some model which computes some score for us. We then can compare this score to our subjective judgement on whether the image is good or bad\footnote{We do not take into account more sophisticated methods for comparison subjective and objective judgements such as mean opinion score (MOS).}. So, first of all, it seems reasonable to stress that any (parametric) model is only restrictively applicable even in the sphere it was designed for. Probably, as it often happens, in some parameter regime the model declines from the phenomenon it endevours to model, and we would inevitably obtain irrelevant results. It does not mean, however, that the model is bad; it merely means it has some sphere of application and has to be used cautiously.

But it seems the problem we tackle dealing with adversarial examples in the field of machine learning is subtler. One argues that using some small deviation from the observed data the model behaves inadequately (provides wrong classification, abnormally high score etc.). This is often expressed by saying that the model is not robust. This is clear for the classification tasks where this can imply that some small perturbation of data changes the class to which the data is assigned. However, the problem we deal with is not that of classification. Instead, we observe gradual changes in VMAF score when increasing the perturbation amplitude. First of all, the natural question here is what the smallness of deviations mean. Specifically, we have to decide in what norm we measure the smallness. Even if the norm is specified (there is a tendency to rely on some euclidean norm, like $l_{\infty}$ or $l_{2}$ as a measure of closeness) one of the problems when dealing with adversarial examples is that image perturbations may not be perceivable with the human eye. In other words, an image can be judged as good subjectively, but after some small changes in terms of the specified norm it is still evaluated as close to the previous one, while its objective score changes significantly. This only may mean that the subjective judgement of what is small and what is big deviation or, to put it another way, the putative subjective norm underlying our judgement is not quite suitable for the model, i.e., is not in good correspondence with typically used norms and that the sensitivity or the resolution power of human visual system (HVS) is not aligned to the model in question. Thus, it seems the question of seeking more suitable norms still remains.

There is also another point to consider. The data presented to the model is implied to be generated by some {\it natural} process. We mean, for example, the image our retinal cells normally obtain is constructed as some natural process in the external environment\footnote{Various forms of animation, though presenting synthetic images, are viewed here as natural-like ones.}. The question then is whether some deviations or perturbations, even small ones in some sense, keep us close to the data {\it likely occurring} in nature (Cf. \cite{GoodfellowSS14}, p. 2). If this is not the case, then the divergence of the model and subjective judgement may be accounted for by the same arguments as provided above. If, on the other hand, naturally occurring data do result in inadequate model behavior, the problem becomes more complicated. But whether we have evidence of {\it naturally occurring} images extremely close to the other ones subjectively and providing noticeably different quality scores?  

The model was trained using some data and the tacit assumption is that this data is {\it typical} or representative for the phenomenon we are trying to model. In the case of VMAF this means that if the goal is to simulate human visual system (HVS) decision on the quality of an image, we need to guarantee that the data we present to the model are those HVS typically deals with.

Having said all this, we would like to stress that we realize that there exists such problem as intentional image alteration to increase the quality score. But the discussion above as well as the approach in the paper looks at this problem only from the point of view of image transformations and perturbations as a natural process. In this sense the generation of perturbations we studied has no intention to be a means to ``break down'' or ``deceive'' but, much more moderately, just investigate numerically the well-established model (VMAF) for image quality assessment from the point of view of bounds within which it remains valid. In this sense, we believe, we move in the direction outlined in the previous discussion: to investigate the properties of the adversarial perturbations resulting in the significant growth of VMAF score. It may be just added that the possibility of re-training the model using adversarial examples is not considered here and would be published elsewhere.

\section*{Conclusions}

In this work we provided the analysis of adversarial perturbations based on a differentiable version of VMAF implemented in the Pytorch framework. Using this implementation we tried to investigate statistical properties of the adversarial perturbations generated by the stochastic gradient algorithm using VMAF as a loss function in terms of $l_{2}$ and $l_{\infty}$ norms. Among several observations made in this work, those worth reiterating here are of weak linear dependence of the perturbations amplitudes on the brightness in the pixels. This observation can be accounted for by edge sharpening effects usually found when increasing VMAF. For this explanation we provided sufficient evidence. On the other hand, this observation can be viewed as a zero-order approximating model of adversarial perturbations. Being purely empirical, this model requires further analysis as a potential means for perturbations generation. We presume to elaborate this further. 

In addition, some observations on the spectra of perturbations also indicate that the adversarial perturbations produced using VMAF keep the image structure on mid- and high-frequency scales. This implies the perturbations affect the image in such a way to keep the smaller objects such as edges.

On the other hand, experiments with image restoration demonstrate different behavior of VMAF sub-metrics for this problem, thus indicating flaws of this quality metric which may be of importance for designing more robust metrics in the future.

\section*{Data availability statement}

The authors declare that the data supporting the findings of this study are available through the links provided in the reference list.


%
\bibliographystyle{IEEEtran}
\bibliography{bibtex/bib/IEEE_IQA_and_metrics,bibtex/bib/IEEE_Image_processing,bibtex/bib/IEEE_Natural_Image_Statistics,bibtex/bib/IEEE_ML}


\section*{Acknowledgment}

The authors are grateful to their colleagues in Algorithm Innovation Lab for discussions.
\ifCLASSOPTIONcaptionsoff
  \newpage
\fi

\end{document}